\documentclass[runningheads]{llncs}

\usepackage{eccv}

\usepackage{eccvabbrv}

\usepackage{graphicx}
\usepackage{booktabs}
\usepackage{multirow}
\usepackage{algorithm}
\usepackage{algpseudocode}
\usepackage{amsmath}
\usepackage{caption}

\usepackage[accsupp]{axessibility}  

\usepackage{hyperref}

\usepackage{orcidlink}

\begin{document}

\title{Efficient and Scalable Monocular Human-Object Interaction Motion Reconstruction} 


\author{
Boran Wen\textsuperscript{1,2}$^*$~~~Ye Lu\textsuperscript{1}$^*$~~~Sirui Wang\textsuperscript{4}~~~Keyan Wan\textsuperscript{3}~~~Jiahong Zhou\textsuperscript{1}~~~
Junxuan Liang\textsuperscript{1,2}\\Xinpeng Liu\textsuperscript{1,2}~~~Bang Xiao\textsuperscript{1}~~~Ruiyang Liu\textsuperscript{5}~~~
Yong-Lu Li\textsuperscript{1,2}$^\dagger$
\\
\small{\textsuperscript{1}SJTU,
\textsuperscript{2}SII,
\textsuperscript{3}FDU,
\textsuperscript{4}BJTU,
\textsuperscript{5}ZJU}
\\
\small{$^*$Equal contribution. $^\dagger$Corresponding author: wenboran@sjtu.edu.cn}
}

\institute{}
\maketitle
\begin{abstract}
Generalized robots must learn from diverse, large-scale human-object interactions (HOI) to operate robustly in the real world. Monocular internet videos offer a nearly limitless and readily available source of data, capturing an unparalleled diversity of human activities, objects, and environments. 
However, accurately and scalably extracting 4D interaction data from these in-the-wild videos remains a significant and unsolved challenge. To overcome the annotation bottleneck, we introduce an efficient sparse contact annotation paradigm. To scale this process, we develop \textbf{InterPoint}, a multi-modal predictor that drives a human-in-the-loop data engine. Building upon these efficiently acquired annotations, we introduce \textbf{4DHOISolver}, a novel optimization framework that constrains the ill-posed 4D HOI reconstruction problem, maintaining high spatio-temporal coherence and physical plausibility.
Leveraging this framework, we introduce \textbf{Open4DHOI}, a new large-scale 4D HOI dataset featuring a diverse catalog of \textbf{135} object types and  \textbf{133} actions.
Furthermore, we demonstrate the effectiveness of our reconstructions by enabling an RL-based agent to imitate the recovered motions. 
\textit{Data and code will be publicly available at \url{https://github.com/wenboran2002/open4dhoi_code}}
\end{abstract}

\begin{figure}
    \centering
    \includegraphics[width=0.999\linewidth]{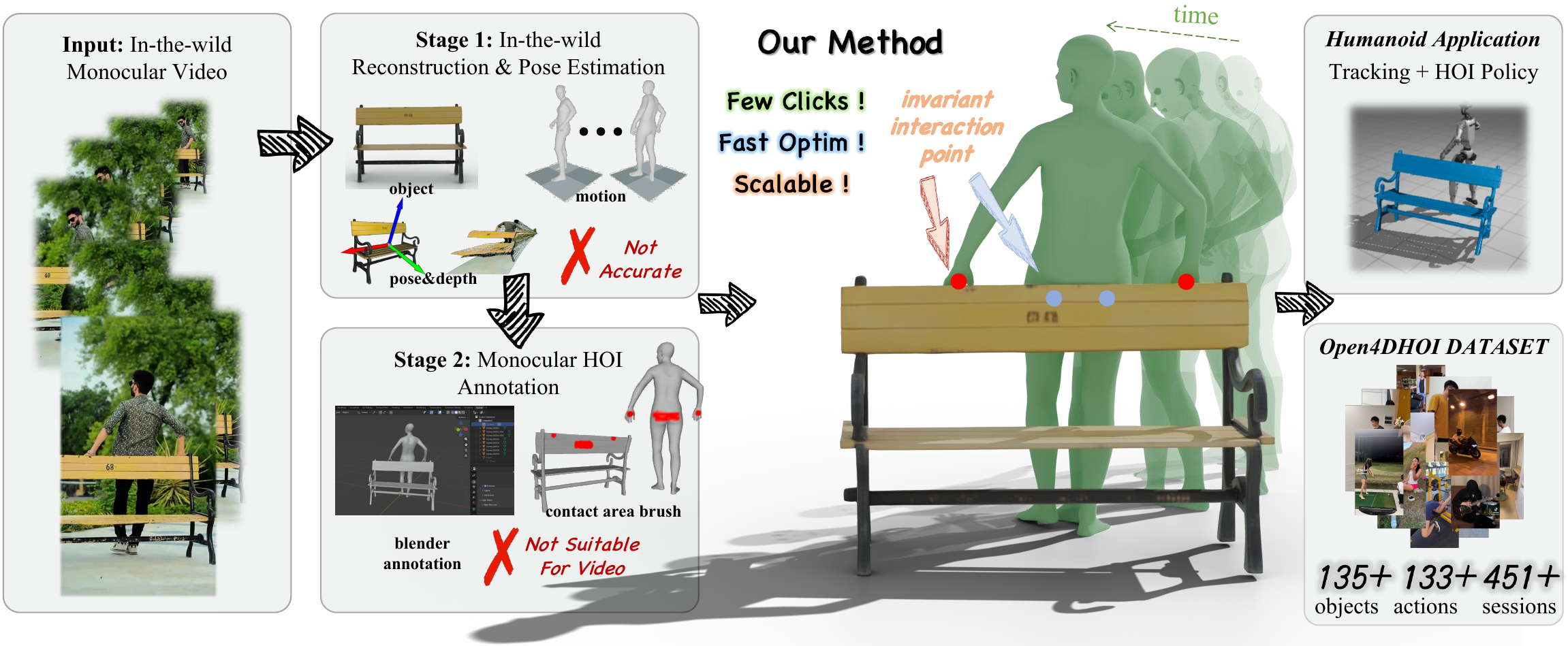} 
    \vspace{-5px}
    \caption{Unlike prior works limited by inaccurate pose/depth alignment or non-scalable single-frame annotations, our method leverages invariant interaction points for fast, video-level 4D HOI reconstruction. This scalable framework enables the efficient construction of our large-scale Open4DHOI dataset and directly facilitates downstream tasks like humanoid learning.
} 
    \vspace{-10px}
    \label{fig:insight}
\end{figure}

\section{Introduction}

The quest for generalized robot systems, capable of understanding and interacting with the complex real world, fundamentally relies on rich, high-fidelity 4D (3D + time) human-object interaction (HOI) data. This data is invaluable for providing crucial insights into human motor skills, intent, and physical reasoning, empowering applications ranging from training intelligent robotic agents for dexterous manipulation to populating virtual worlds with realistic digital humans for gaming and VR/AR applications.

Despite its importance, acquiring generalized 4D HOI data remains a significant challenge. Existing multi-sensor oriented HOI capture systems, exemplified by BEHAVE~\cite{behave}, offer high-precision reconstructions by leveraging multi-view camera setups and sophisticated sensor arrays. However, the associated cost, from numerous cameras, sensors, and specialized studios, is prohibitive for large-scale data collection. Consequently, these systems are often confined to controlled indoor settings with limited object diversity, preventing the capture of HOI data for common outdoor activities like riding and surfing, as well as complex on-site construction tasks involved in industrial manufacturing processes.

Recognizing the limitations of high-precision capture systems, recent research has explored reconstructing 3D HOI from widely available monocular images/videos. Since purely automated pose and depth alignment often yields physically inaccurate contacts~\cite{sam3dobjects, sam3dbody, gio}, methods like Open3DHOI~\cite{open3dhoi} rely on manual object pose adjustment in Blender, while PICO~\cite{pico} proposes annotating contact regions. However, while effective for static images, both annotation strategies become prohibitively expensive and time-consuming when applied to video sequences, and they fundamentally struggle to enforce spatio-temporal consistency across frames. To address these critical challenges, we propose \textbf{4DHOISolver}, a novel and efficient framework for generating high-quality, temporally-consistent 4D HOI data from diverse monocular internet videos.

Our core idea is to replace expensive, dense per-frame labeling with a lightweight annotation of temporally invariant interaction points, guided by system-provided interactable reference points on the human body parts. To scale up this process, we introduce a human-in-the-loop (HITL) data engine powered by a multi-modal contact predictor, InterPoint. InterPoint automatically proposes initial annotations from monocular frames, reducing manual effort. More importantly, this establishes a positive feedback loop: as more data is verified and annotated by humans, InterPoint is continuously fine-tuned, growing progressively more accurate and thereby accelerating the annotation of increasingly complex interactions. Then, we develop \textbf{4DHOISolver}, a two-stage framework that first performs a rapid geometric alignment using least-squares matching and inverse kinematics, followed by a gradient-based optimization to refine the interaction's physical plausibility. 
Leveraging this pipeline, we construct \textbf{Open4DHOI}, a diverse dataset containing 451 videos across 135 object categories and 133 actions. Furthermore, we validate the potential of our data for downstream robotic applications by designing a novel contact-guided reward function, enabling an RL-based agent to master complex HOI motion imitation.
Overall, our work includes the following contributions:
\begin{itemize}

    \item We propose a scalable human-in-the-loop data engine driven by a contact point predictor (InterPoint). It automatically proposes human-object contact pairs and continuously improves through a data flywheel, reducing annotation costs.

    \item We propose \textbf{4DHOISolver}, a novel framework that reconstructs high-fidelity, physically plausible, and spatio-temporally coherent 4D HOI from monocular video by constraining a two-stage optimization with sparse contact point annotations.
    
    \item Leveraging this pipeline, we build and release \textbf{Open4DHOI}, a new, large-scale 4D HOI dataset.
    
    \item We demonstrate the effectiveness and utility of our dataset by developing a novel, contact-guided reward function and successfully training an RL-based agent to perform challenging HOI motion imitation.

\end{itemize}

\section{Related Works}

\subsection{3D/4D HOI Datasets}
Traditional 3D/4D HOI datasets heavily rely on constrained studio environments and complex hardware. Capture setups range from calibrated multi-view RGB-D systems~\cite{behave, intercap, core4d} and massive camera arrays~\cite{neuraldome, hoim3} to wearable Mocap suits~\cite{omomo, humoto, trumans}. Consequently, object tracking in these datasets often requires labor-intensive or intrusive methods, such as manual annotation~\cite{behave}, physical markers~\cite{omomo, trumans, choice}, or sophisticated multi-camera trackers~\cite{humoto, foundationpose}.

To overcome these restrictive capture conditions, recent efforts~\cite{wildhoi, open3dhoi, pico, lemon, interactvlm, cari4d, dynhor, scorehoi} have shifted towards reconstructing HOI from in-the-wild monocular videos. While human motion is typically recovered using off-the-shelf pose estimators, handling object motion remains diverse and challenging. Current object-centric strategies include optimizing pre-defined template keypoints~\cite{wildhoi}, manually refining generated 3D meshes in Blender~\cite{open3dhoi}, or retrieving proxy meshes from large databases via contact constraints~\cite{pico}.

\subsection{3D Reconstruction Tools}
Robust 3D reconstruction of humans and objects is fundamental to recovering HOI from monocular videos. For human motion, while methods~\cite{osx, smplerx, aios}, offer unified body-hand-face recovery, they often lack global trajectory awareness. Therefore, we leverage human motion recovery methods~\cite{gvhmr, tram, wham, slahmr, 4dhumans} to accurately reconstruct global body motion aligned with the camera space. To compensate for the lack of hands, we integrate with hand reconstruction methods~\cite{hamer, wilor, hamr, handos, dynhamr}, achieving high-fidelity, full-body kinematics essential for interaction analysis.

For object reconstruction, recent 3D generative models~\cite{trellis, trellis2, shi2023zero123plus, xu2024instantmesh, fancy123} trained on massive datasets~\cite{objaverseXL} can now produce high-quality meshes from real-world images. Furthermore, methods targeting real ``in-the-wild'' images, such as SAM 3D~\cite{sam3dobjects}, enable object 3D reconstruction even in the presence of occlusions.

\subsection{Contact and Affordance Prediction}
Previous works often treat HOI contact prediction by isolating the human and object domains, such as predicting dense human contact maps~\cite{deco, rich} or localizing object affordances~\cite{3daffordancellm, laso, iagnet}. However, this treatment neglects the coupled nature of physical interactions~\cite{interprior, xu2025intermimic}. While recent methods like InteractVLM~\cite{interactvlm, lemon} bridge this gap by leveraging VLMs to jointly predict continuous bilateral contacts, generating these dense maps remains computationally expensive~\cite{xue2025rog}. Building upon this joint-prediction paradigm, we propose transitioning from dense continuous maps to highly efficient, sparse discrete contact points, specifically designed to power a scalable human-in-the-loop annotation engine.

\section{Pipeline}
In this section, we introduce how we efficiently collect and annotate in-the-wild, open-vocabulary HOI data, as well as perform fast reconstruction based on these annotations. Our core idea is to annotate temporally invariant human–object interaction ``point pairs'' and perform fast optimization-based reconstruction through these persistent point correspondences.





Specifically, in Sec.~\ref{sec:4d_reconstruction}, we introduce a coarse initialization reconstruction method using existing 3D reconstruction techniques. Sec.~\ref{sec:annotation_app} presents the efficient and scalable annotation method and our app. In Sec.~\ref{sec:interpoint}, we propose an interaction ``point-pair'' prediction model. By predicting invariant interactive points prior to manual annotation, this approach effectively alleviates the annotation burden and facilitates a human-in-the-loop annotation paradigm. In Sec.~\ref{sec:4dhoisolver}, we describe how to achieve efficient and high-quality reconstruction from the annotated point pairs. Finally, in Sec.~\ref{sec:dataset}, we introduce our constructed dataset Open4DHOI with various annotations and diversity.

\subsection{4D Reconstruction}
\label{sec:4d_reconstruction}

\begin{figure}[t]
  \centering
  \begin{minipage}{0.48\textwidth}
    \centering
    \includegraphics[width=\linewidth]{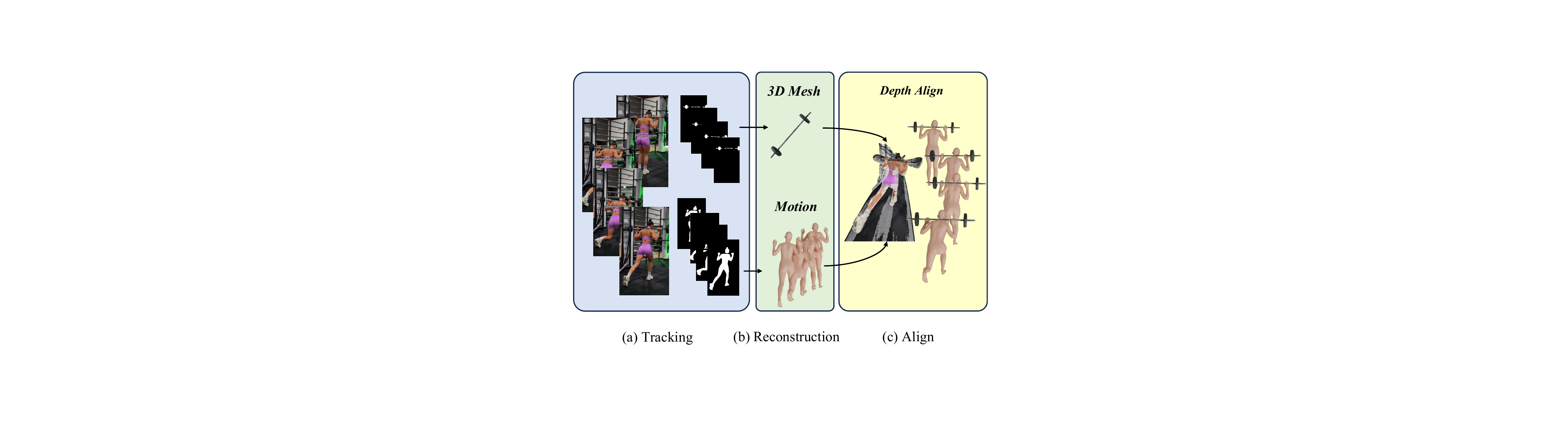}
    \caption{Our automated 4D reconstruction pipeline consists of three components: (a) human and object tracking, (b) 3D reconstruction, and (c) spatial alignment.}
    \label{fig:auto-recon}
  \end{minipage}\hfill 
  \begin{minipage}{0.48\textwidth}
    \centering
    \includegraphics[width=\linewidth]{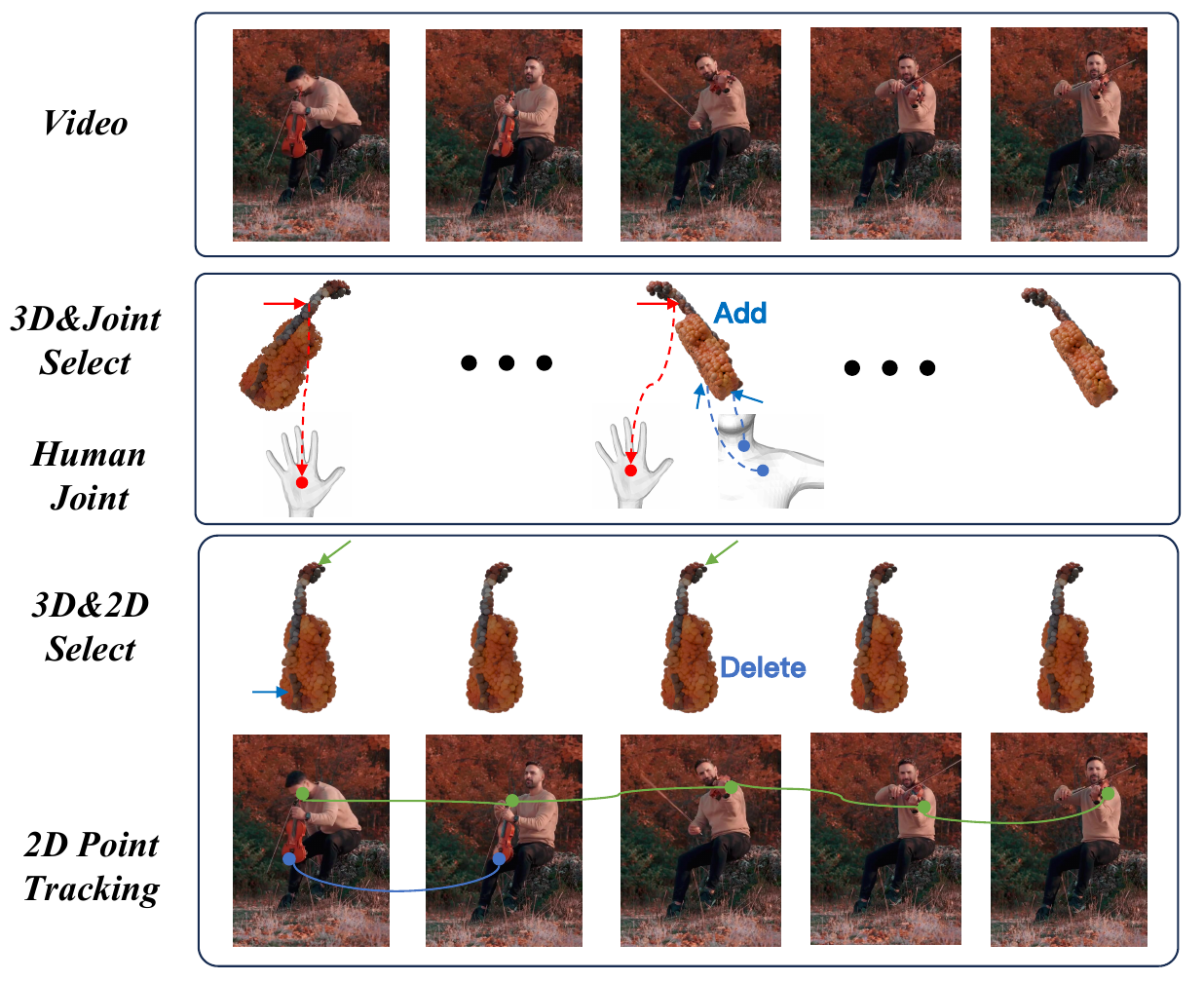}
    \caption{\textbf{Annotation app}: the first row shows the reference video, the second row displays the 3D-Human Joint annotations, and the third row presents the 3D-2D Projection annotations.}
    \label{fig:app}
  \end{minipage}
\end{figure}

To annotate in-the-wild HOI data, we need to leverage state-of-the-art 3D reconstruction tools for initialization. High-quality 3D initialization ensures both annotation efficiency and data quality.
To this end, we propose a robust 4D reconstruction pipeline shown in Fig.~\ref{fig:auto-recon}. Our proposed pipeline consists of three stages: 

1) We preprocess data by detecting video shot transitions, removing non-interactive frames, and tracking masks for humans and objects.

2) We utilize SAM 3D Objects~\cite{sam3dobjects} and GVHMR~\cite{gvhmr} to reconstruct objects and human motions, respectively. Furthermore, we independently reconstruct and integrate the hands to ensure consistency with the overall body pose. 


3) We initialize and align the HOI poses through a depth-aware projection approach. Specifically, we estimate the depth of humans and objects based on DepthAnythingV2~\cite{depth_anything_v2} and use the depth point cloud for the spatial alignment and object scale estimation by projecting the human mesh and object mesh together to fit the point cloud.

\subsection{Annotation App}
\label{sec:annotation_app}
\subsubsection{Contact Definition.}
\label{sec: contact defination}
A key prerequisite for annotating HOI point pairs is to define the interaction keypoints. Previous work predefined keypoints on object templates, which is not suitable for open-world object reconstruction. We choose to divide human joints as finely as possible while freely selecting the corresponding points on the object. We adopt a \textit{tree structure} to define human keypoints, where \textbf{26} main body parts serve as parent nodes, and the child nodes represent finer subdivisions of each part, a total of \textbf{87} keypoints. For example, we divide the forearm into four points—front, back, left, and right—to ensure that contacts from all directions have corresponding interaction points. 

\subsubsection{Annotation Procedure.}
To annotate HOI motion efficiently, we built an annotation app, as shown in Fig.~\ref{fig:app}. The user needs to annotate two parts. First, annotating the contact point pair between the object and the human body by selecting a 3D point $p_i^{3D}$ on the point cloud and the corresponding human joint $q_i^{3D}$ in the joint tree. Second, constraining the object pose according to the video by annotating the 3D object point $p_i^{2D}$ and clicking the corresponding 2D point $q_i^{2D}$ on the frame.

Based on the fact that there are usually some fixed points in HOI motion, we aim to find and track these points, which makes the annotation process easy and fast. The user only needs to re-annotate when the stable points are changed in the video. Specifically, we use the Point Tracking model~\cite{cotracker} to track the annotated 2D points. Interactive objects are divided into movable and static categories. For objects that remain motionless throughout the video, we provide an additional annotation option, allowing users to indicate whether the object pose should be fixed based on the video content. What's more, for cases in Sec.~\ref{sec:4d_reconstruction} where the reconstructed object scale is inaccurate, we adjust the object’s scale before annotation.

\begin{figure*}[t]
    \centering
    \includegraphics[width=0.90\linewidth]{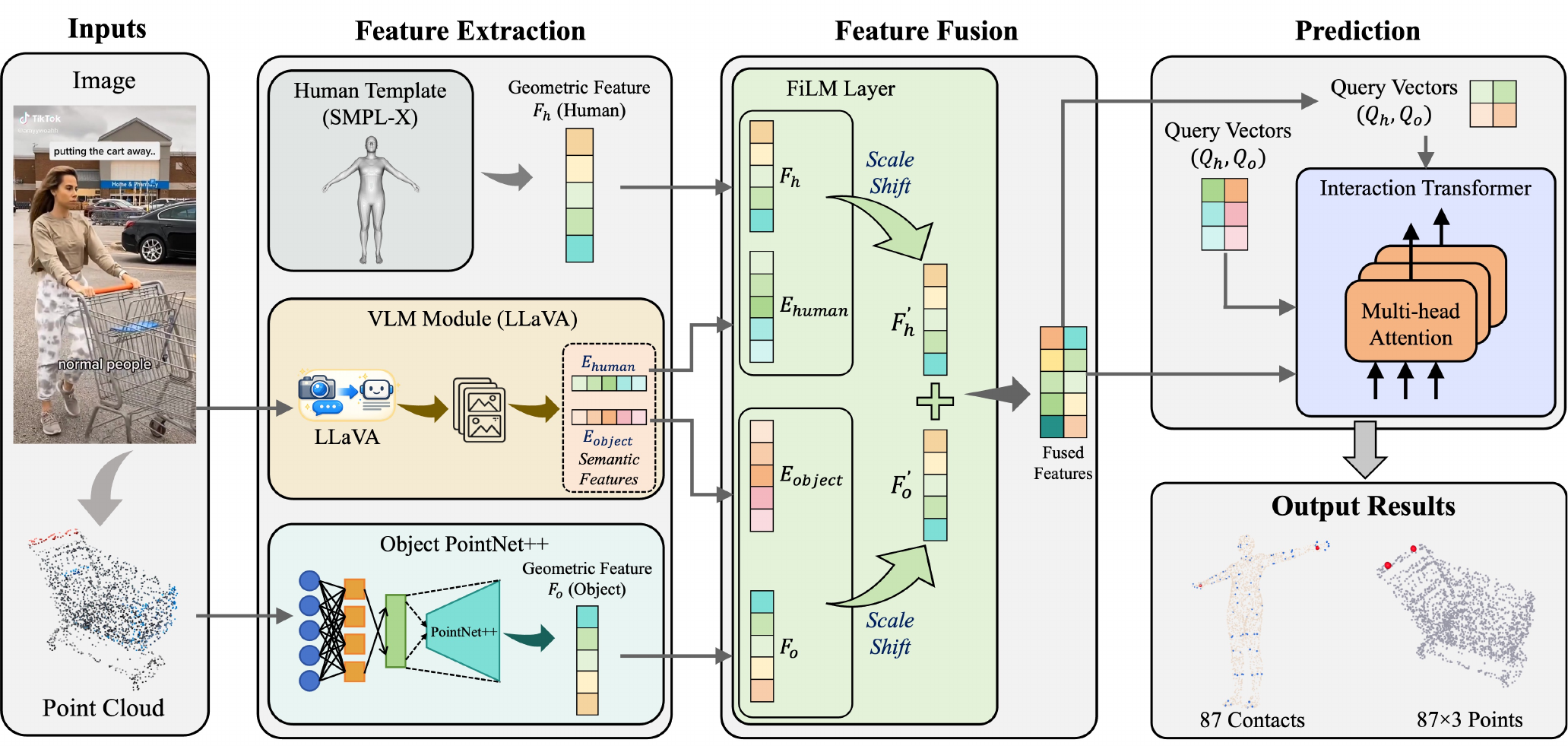} 
    \caption{Network architecture.} 
    \vspace{-10px}
    \label{fig:network}
\end{figure*}
\subsection{InterPoint Model}
\label{sec:interpoint}
To further accelerate annotation, we introduce \textbf{InterPoint} to initialize annotations. Annotators can obtain high-quality annotations with minor refinements based on initial predictions, substantially reducing data collection cost. Meanwhile, newly acquired annotations are fed back to train InterPoint, yielding better initializations over time and forming a data flywheel that enables rapid scaling of the dataset. 
\subsubsection{Problem Formulation.}
The training objective of our model is to predict the interactive points $p^{3d}$ and $q^{3d}$ for a given frame. Given this frame and a 3D object (downsampling to 1024 points) as inputs, the model outputs the interacting human keypoints $q^{3d}$—identified from a predefined set of 87 body keypoints—along with their corresponding contact points $p^{3d}$ on the object surface.

\subsubsection{Model Architecture.}
Our InterPoint model extracts 2D semantic embeddings ($E_h, E_o$) via a VLM~\cite{llava} and 3D object features $F_o$ using a PointNet++~\cite{pointnet++} encoder. Since all humans share a common SMPL-X body model, we introduce learnable template parameters to form the 3D human feature $F_h$. To bridge 2D semantics and 3D geometry, a Feature-wise Linear Modulation (FiLM) layer dynamically injects $E_h$ and $E_o$ into the 3D point features $F_h$ and $F_o$. For deep feature fusion and interaction modeling, an Interaction Transformer concatenates these enriched human and object representations into a unified memory space. Finally, learnable queries ($Q_h, Q_o$) attend to this shared memory to predict 87 human joint contact probabilities and localize object contact coordinates via an attention mechanism. The architecture of our model is shown in Fig.~\ref{fig:network}.

\vspace{1mm}\noindent\textbf{Training Recipe.} 
We supervise the human keypoint predictions $\hat{q}_{3D} \in \mathbb{R}^{87}$ via a standard BCE loss ($\mathcal{L}_{\text{human}}$). However, for object-side correspondences, the extreme sparsity (only 2--3 positive entries among $87 \times 1024$ pairs) causes vanilla BCE to collapse into trivial solutions. To address this, we formulate a bidirectional contrastive loss.

For each ground-truth human keypoint $k \in K$, we define its positive object point set $\mathcal{P}_{k}$ using spatial proximity (e.g., radius constraints and KNN) to the object point cloud. Let $s_{k,n}$ be the predicted correspondence score. The keypoint-to-object loss encourages $k$ to assign high probability mass to its geometric neighborhood $\mathcal{P}_{k}$:
\begin{equation}
    \mathcal{L}_{k\to o} = -\frac{1}{|K|} \sum_{k\in K} \log \frac{\sum_{n\in\mathcal{P}_k}\exp(s_{k,n}/\tau)}{\sum_{n=1}^{N}\exp(s_{k,n}/\tau)},
\end{equation}
where $\tau$ is the temperature. Conversely, for each object point $n$, its positive keypoint set is $\mathcal{Q}_n = \{\, k \mid n \in \mathcal{P}_k \,\}$. The object-to-keypoint loss $\mathcal{L}_{o\to k}$ is symmetrically defined over valid object points $\mathcal{N}_+ = \{\, n \mid \mathcal{Q}_n \neq \varnothing \,\}$:
\begin{equation}
    \mathcal{L}_{o\to k} = -\frac{1}{|\mathcal{N}_+|} \sum_{n\in\mathcal{N}_+} \log \frac{\sum_{k\in\mathcal{Q}_n}\exp(s_{k,n}/\tau)}{\sum_{k=1}^{K}\exp(s_{k,n}/\tau)}.
\end{equation}
The final objective is computed as $\mathcal{L}=\lambda_h\mathcal{L}_{\text{human}} +\lambda_{k2o}\mathcal{L}_{k\to o} +\lambda_{o2k}\mathcal{L}_{o\to k}$. We train InterPoint on our Open4DHOI dataset (Sec.~\ref{sec:dataset}), detailing results in Sec.~\ref{sec:interpoint_vlm}.

\subsection{4DHOISolver}
\label{sec:4dhoisolver}
After obtaining the point annotations, we adopt a two-stage reconstruction approach called \textbf{4DHOISolver} in Fig.~\ref{fig:hoisolver}. 
In the first stage, we perform fast point-pair matching based on least squares and apply rapid inverse kinematics (IK) optimization to adjust the human limb positions. 
In the second stage, we refine the interaction’s physical plausibility through gradient-based optimization. 
This two-stage optimization framework ensures high efficiency while maintaining reconstruction accuracy.

\subsubsection{HOI Keypoint Solver.}
To achieve faster optimization while ensuring accuracy, we further designed a point matching method to align keypoint pairs and optimize the pose of the object. After completing the object pose optimization, we refine the human limb positions using IK.

Our core algorithm uses the least squares method to solve two Points-Alignment problems: 3D-3D spatial alignment and the 3D-2D projection alignment, as shown in Eq.~\ref{equ:point-alignment}, where the optimization targets are $\mathbf{R}_o$ and $\mathbf{t}_o$. 

\begin{equation}
\begin{aligned}
\mathbf{r}^{3D}_i(\boldsymbol{\theta}) &= \sqrt{w_{3D}}\left( \mathbf{R}_o\,\mathbf{p}^{3D}_i + \mathbf{t}_o - \mathbf{q}^{3D}_i \right), \\
\mathbf{r}^{2D}_j(\boldsymbol{\theta}) &= \sqrt{w_{2D}}\left( \pi_{\mathbf{K}}\left(\mathbf{R}_o\,\mathbf{p}^{2D}_j + \mathbf{t}_o\right) - \mathbf{q}^{2D}_j \right), \label{equ:point-alignment} \\
\boldsymbol{\theta}^* &= \arg\min_{\boldsymbol{\theta}} \frac{1}{2}\left( \sum_{i=1}^{N} \left\| \mathbf{r}^{3D}_i(\boldsymbol{\theta}) \right\|_2^2 + \sum_{j=1}^{M} \left\| \mathbf{r}^{2D}_j(\boldsymbol{\theta}) \right\|_2^2 \right).
\end{aligned}
\end{equation}

By aligning the annotated point pairs, we can track the object's pose and spatial position, as well as achieve coarse alignment of the contact regions.
Since the limbs are the most critical parts for interaction and are prone to spatial misalignment, we apply a separate IK-based quick optimization specifically for the limbs.


\subsubsection{HOI Optimizer.} Building on the initialization provided by the HOI solver, we propose an HOI optimizer to further refine the physical properties of Human–Object Interactions by optimizing ${ \theta}_h$, $\mathbf{R}_o$, and $\mathbf{t}_o$. 

To make the optimization focus more on the annotated contact regions while maintaining stability in the non-contact areas, we propagate gradients only along the kinematic chain based on our joint-tree distances, ensuring that joints not directly involved in the optimization remain unaffected.

{\textbf{Loss Function}.}
Our optimization process is driven by a composite loss function consisting of three terms: contact loss, collision loss, and mask loss, as shown in Eq.~\ref{eq:opt loss}.
\begin{equation}
\begin{aligned}
    L = w_c L_{\text{contact}} + w_{coll} L_{\text{collision}} + w_{m} L_{\text{mask}}.
\end{aligned}
\label{eq:opt loss}
\end{equation}

$L$ addresses the limitation of HOI Keypoint Solver~\cite{hoikeypointsolver}’s point-based optimization, which ignores the physical plausibility of interaction, such as penetration, and allows for more fine-grained adjustments of the object’s pose estimation.

{\textbf{Post-Smoothing.}}
To ensure optimization efficiency, we perform optimization every $K$ frames and interpolate the intermediate frames while applying a low-pass filter to smooth the entire optimized motion.

The overall optimization pipeline is summarized in the Alg.~\ref{alg:hoi_optimization}.

\begin{algorithm}[t]
\caption{4DHOISolver Optimization Pipeline}
\label{alg:hoi_optimization}
\begin{algorithmic}[1]
\Require Video frames with annotated 3D-3D and 3D-2D correspondences
\State Initialize model parameters (body pose, shape, object pose)
\For{each frame $t$ with step size K}
    \State Perform weighted least squares optimization to estimate object pose $(\mathbf{R}_t, \mathbf{t}_t)$ using 3D-3D and 3D-2D correspondences
    \State Run IK to refine human body pose parameters based on 3D keypoints
    \For{$i = 1$ to $20$}
        \State Optimize body and object poses with collision, mask, and contact losses using the \textbf{Adam optimizer}
    \EndFor
    \If{$t > 0$}
        \State Interpolate parameters linearly between frames $t-K$ and $t$
    \EndIf
\EndFor
\State Apply low-pass filter smoothing on optimized poses across all frames
\State \Return Optimized human and object poses
\end{algorithmic}
\end{algorithm}

\begin{figure*}[t]
    \centering
    \includegraphics[width=0.9\linewidth]{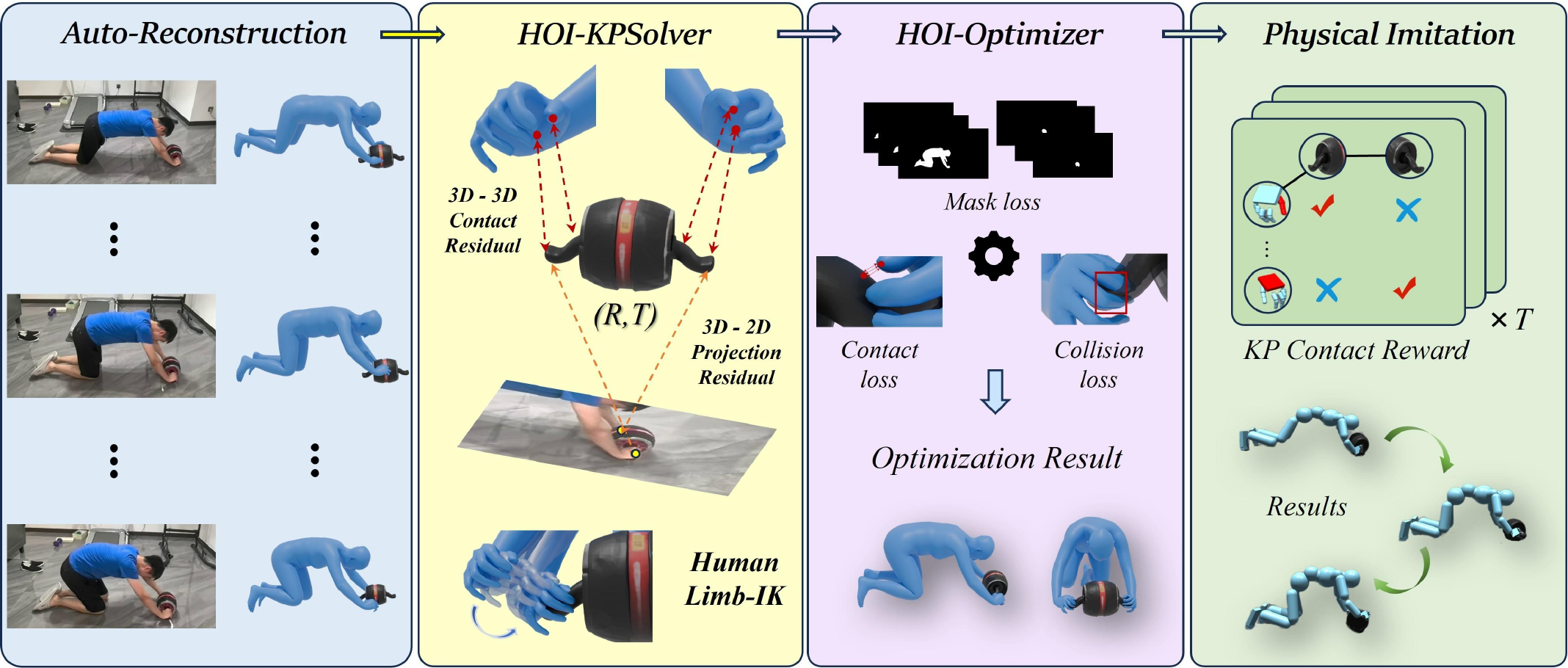} 
    \vspace{-5px}
    \caption{\textbf{Pipeline:} Our reconstruction pipeline consists of four stages. First, we perform automated reconstruction as described in Sec.~\ref{sec:4d_reconstruction}. After obtaining the reconstructed results, we apply the 4DHOISolver from Sec.~\ref{sec:4dhoisolver} for optimization based on the annotations. Finally, we conduct physical imitation as described in Sec.~\ref{sec:hoi_simulation}.
} 
    \vspace{-10px}
    \label{fig:hoisolver}
\end{figure*}

\subsection{Dataset}
\label{sec:dataset}
We collected a dataset called \textbf{Open4DHOI} with \textbf{451} sequences comprising \textbf{131k} frames across \textbf{135} categories of rigid objects via two sources: mobile-phone capture and TikTok crawling. 
Among them, \textbf{299} sequences (\textbf{79k} frames) were self-collected using mobile phones, and \textbf{152} sequences (\textbf{52k} frames) were crawled from TikTok.

\begin{figure*}
    \centering
    \includegraphics[width=0.9\linewidth]{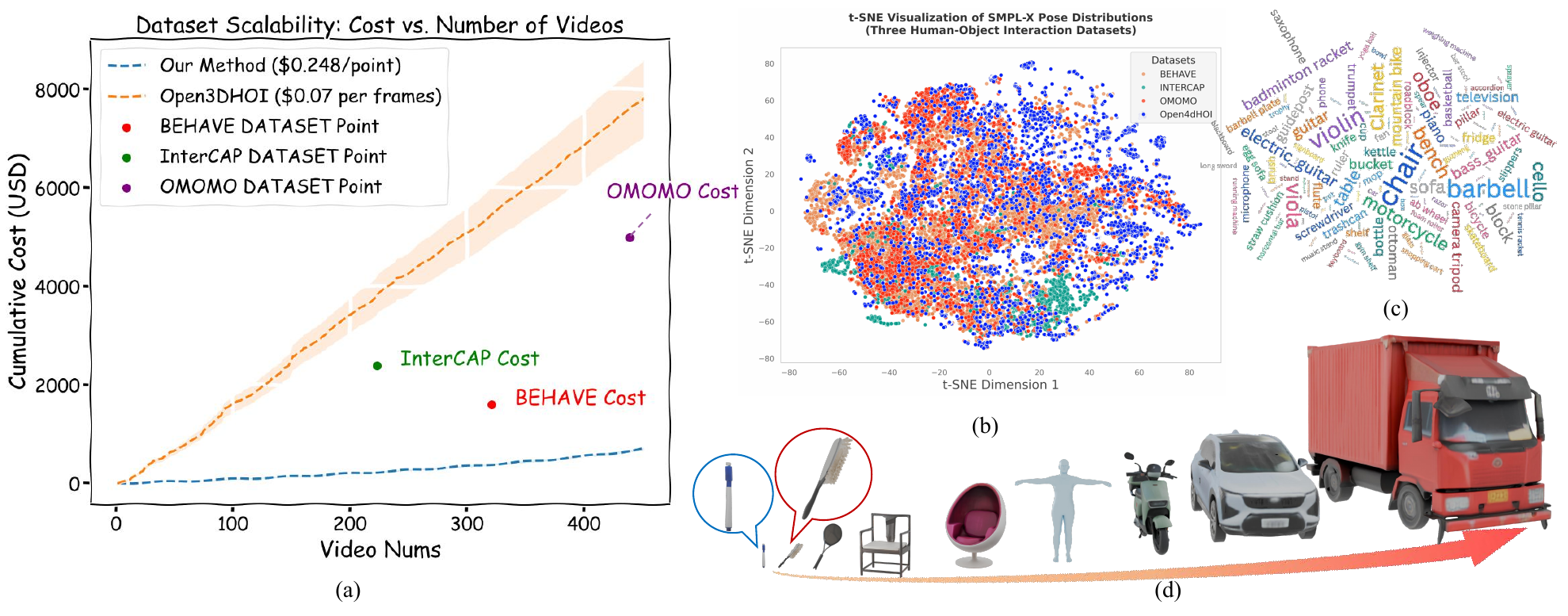} 
    \vspace{-5px}
    \caption{Dataset characteristics: (a) demonstrates that our dataset can be scaled up massively at minimal cost, (b) illustrates comparison of body pose diversity across different datasets, (c) presents a word cloud of the objects, and (d) depicts a scale-based visualization of the objects from smallest to largest.
} 
    \vspace{-10px}
    \label{fig:dataset}
\end{figure*}

It is worth noting that our data collection process is highly cost-efficient and requires no additional capture equipment. Our data annotation process takes approximately ten minutes per video and involves a simple, user-friendly workflow. In contrast, a single Azure Kinect RGB-D camera costs around $\$399$, while a complete Vicon motion capture system is priced at approximately $\$50,000$. On average, each of our videos requires annotation of 6.24 points. Based on the Amazon Mechanical Turk (AMT) hourly wage, the cost curve is plotted in Fig.~\ref{fig:dataset} (b). Meanwhile, we compare the costs and scales of Open3DHOI using Blender for frame-by-frame annotation with other 4D HOI datasets. It can be seen that our annotation method has strong scalability.

\section{HOI Simulation}
\label{sec:hoi_simulation}

To demonstrate the scalability of our data and its applicability to downstream tasks such as humanoid robotics, we train a policy $\pi$ via reinforcement learning to control a humanoid to imitate our reconstructed HOI motions $\{q_t^h, q_t^o\}_{t=1}^T$. Since monocular reconstructions inherently contain physical artifacts (e.g., interpenetration, floating) under unseen viewpoints, existing rigid tracking methods~\cite{Luo2023PerpetualHC,Wang_2025_CVPR,wang2023physhoiphysicsbasedimitationdynamic} struggle. We therefore propose a noise-robust simulation based on InterMimic~\cite{xu2025intermimic}, incorporating our contact annotations as strong priors. The state comprises human joint transforms $q_{t}^h \in \mathbb{R}^{52 \times 6}$ and object poses $q_{t}^o \in SE(3)$.

\noindent\textbf{Tracking Reward.} 
To encourage basic motion imitation, we use a standard tracking reward $R_{\text{tr}} = \exp( -( E_{\text{p}} + E_{\text{v}} ) )$. The position error $E_{\text{p}}$ computes the weighted squared differences in joint positions $p$ and rotations $\theta$ (via $\ominus$) for both the human and object compared to the reference. Similarly, $E_{\text{v}}$ penalizes differences in linear ($v$) and angular ($\omega$) velocities.






\noindent\textbf{KP Label Reward}.
To constrain interaction regions, we map the 87 predefined human keypoints from 4DHOISolver to the 52 humanoid joints. We use the annotated keypoint pairs as ground-truth (GT) contact labels $\hat{c}$ and treat the simulated force feedback as predicted labels $c$. The discrepancy is penalized specifically for active GT contacts:
\begin{equation}
    R_{lr} = \sum \|\hat{c}-c\| \odot \hat{c},
\end{equation}
where $\odot$ ensures the policy is penalized only when it misses annotated interactions.

\noindent\textbf{KP Contact Reward}.
To compensate for reconstruction inaccuracies, we design a 3D contact reward leveraging our fine-grained annotations. Crucially, we construct an Interaction Graph ($I_g$) to explicitly model the dynamic contact map across frames (Fig.~\ref{fig:hoisolver}). During training, annotated object points $M^o$ move with the object's rigid transformation. We enforce the global spatial distance between paired active contact points in $I_g$ to be zero:
\begin{equation}
    R_{cr} = -\lambda_{c}\sum \left\| \mathbf{R}_t^o\,M^o(I_g) + p_t^o - p_t^h(I_g) \right\|^2,
\end{equation}
where $\mathbf{R}_t^o \in SO(3)$ and $p_t^o \in \mathbb{R}^3$ denote the object's global rotation and translation at frame $t$, and $p_t^h(I_g)$ is the corresponding human joint position.



\begin{figure}[t]
  \centering
  
  \begin{minipage}[t]{0.44\textwidth}
    \vspace{0pt}
    \vspace*{0.4cm} 
    \centering
    \includegraphics[width=\linewidth]{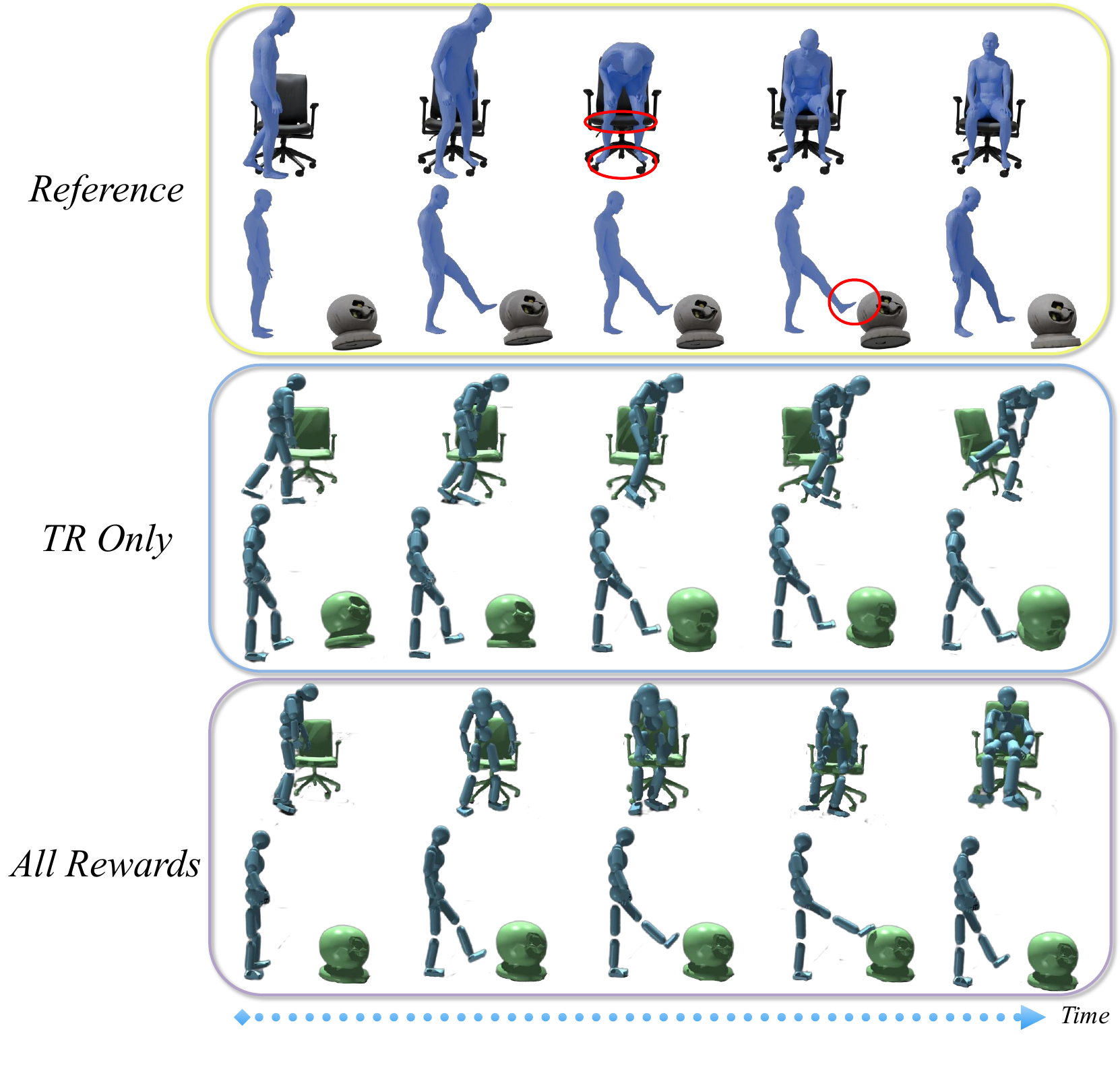}
    \caption{Visualization of our HOI imitation results.}
    \label{fig:simulation}
  \end{minipage}\hfill
  \begin{minipage}[t]{0.52\textwidth}
    \vspace{0pt}
    \centering
    \captionof{table}{HOI simulation results. To assess the performance of our method, we sample \textbf{80} sequences from Open4DHOI, containing a total of \textbf{29k} frames, and group them by action type. Each action is trained under three different reward settings. We evaluate the training results using three metrics: MPJPE (mean per joint position error) with reference motion, contact score, and jitter score to evaluate the smoothness of the motion.}
    \label{tab:simulation}
    \resizebox{\linewidth}{!}{%
      \begin{tabular}{lccc}
        \toprule
        Reward & MPJPE (mm) $\downarrow$ & contact (mm) $\downarrow$ & jitter$\downarrow$  \\
        \midrule
        TR only & 151.82 & 43.94 & 91.43 \\
        TR+LR & 156.54 & 39.53 & 80.37 \\
        TR+LR+CR & \textbf{125.76} & \textbf{26.76} & \textbf{79.63} \\
        \bottomrule
      \end{tabular}
    }
  \end{minipage}
  
\end{figure}

\noindent\textbf{Optimization Results}.
The results of the ablation study are summarized in Tab.~\ref{tab:simulation}. In Fig.~\ref{fig:simulation}, we visualize the imitation results of our method. It can be observed that our approach effectively optimizes physically implausible issues in the reference motion, such as interpenetration, foot floating, and contact misalignment.

\begin{table*}[t]
\centering
\small
\caption{Dataset comparison across motion, object, and interaction diversity.}
\label{tab:dataset_summary}
\setlength{\tabcolsep}{6pt}
\resizebox{0.9\textwidth}{!}{\begin{tabular}{l cc c cc cc cc}
\toprule
\multirow{2}{*}{Dataset} & \multicolumn{2}{c}{Quantity} & \multicolumn{1}{c}{Motion} & \multicolumn{2}{c}{Object} & \multicolumn{2}{c}{Interaction} & \multicolumn{2}{c}{Video} \\
\cmidrule(lr){2-3}\cmidrule(lr){4-4}\cmidrule(lr){5-6}\cmidrule(lr){7-8}\cmidrule(lr){9-10}
 & Frames & Seq. & Div. $\uparrow$ & \#Cat. & Scale & Contact & Action & IS $\uparrow$ & Scene \\
\midrule
BEHAVE     & 15k   & 321   & 2.492 & 10  & $0.16 \sim 0.82$ & B     & 20  & 1.53 & Indoor \\
OMOMO      & 810k  & 4.4k  & 1.022 & 15  & $0.24 \sim 1.70$ &  B     & 34  & -    & Indoor \\
InterCap   & 67k   & 223   & - & 6   & $0.10 \sim 0.69$ & B+H& -   & 2.43 & Indoor \\
IMHD$^2$   & 892k &  295 &  2.497   & 10 & $0.08 \sim 0.36$ & B+H & 82 & 1.82 & Indoor \\
\midrule
PICO       & 4.1k  & -     &  -   & 44  & $0.12 \sim 3.21$ & B+H& -   & -    & Wild   \\
Open3DHOI  & 2.5k  & -     &  -   &133  & $0.02 \sim 5.63$ & B+H&120  & -    & Wild   \\
\midrule
Open4DHOI  &131k   & 451   & 3.465 &135  & $0.02 \sim 4.22$ & B+H&133  & 6.32 & Wild   \\
\bottomrule
\end{tabular}}
\end{table*}

\section{Experiments}

\subsection{Dataset Experiments}
\label{diversity}
Our dataset demonstrates high diversity across multiple dimensions. The objects exhibit a wide range of properties, encompassing various rigid bodies commonly found in everyday life. Human motions are equally diverse, involving movements that engage all major joints. Our dataset is also highly diverse at the interaction level, encompassing a wide range of actions as well as many uncommon HOIs.

{\textbf{Object Diversity}}. A key characteristic of our dataset is the wide variety of object categories. The objects span a wide range of sizes, as shown in the Tab.~\ref{tab:dataset_summary} and Fig.~\ref{fig:dataset} (e).

{\textbf{Motion Diversity}}. Our data is collected from a large number of real-world TikTok videos. It covers a broader range of motions. We adopt the diversity evaluation protocol of MoMask~\cite{guo2023momaskgenerativemaskedmodeling}: we encode human motions into its latent space, compute the pairwise L2 distances between embeddings, and use the mean distance as the diversity metric. The results are reported in Tab.~\ref{tab:dataset_summary}.

{\textbf{Interaction Diversity}}. We used the Qwen2.5-VL-72B~\cite{qwen2.5vl} model to annotate actions in the videos and further manually verified them. Ultimately, our dataset contains 133 action categories. For the BEHAVE~\cite{behave} and OMOMO~\cite{omomo} datasets, we used Qwen2.5-72B~\cite{qwen2.5} to extract action categories from the textual descriptions in the dataset and manually filtered the results, with details in Tab.~\ref{tab:dataset_summary}.


\subsection{4DHOISolver Experiments}

\begin{figure*}[t]
    \centering
    \includegraphics[width=0.73\linewidth]{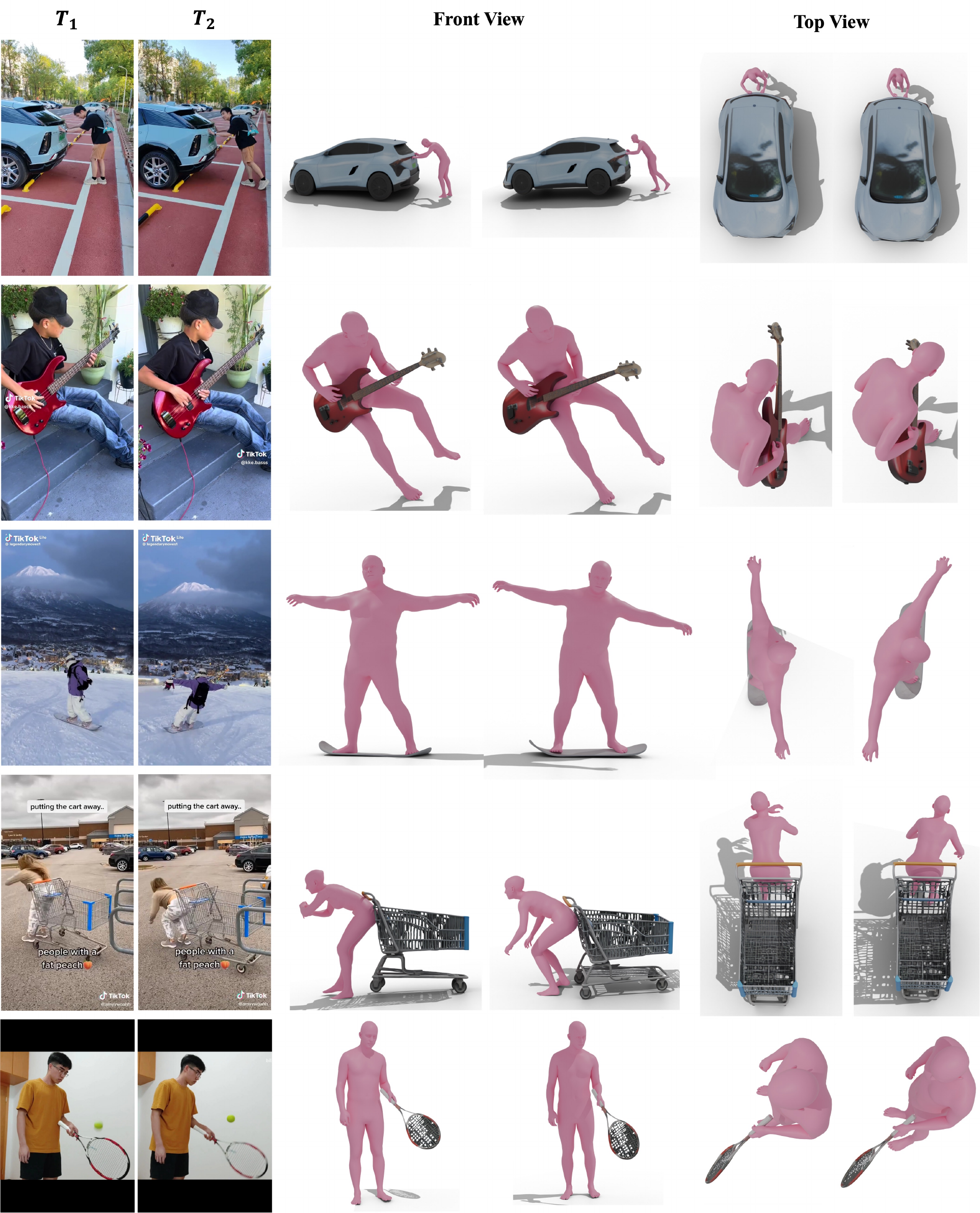} 
    \caption{Visualization of our 4DHOISolver reconstruction results.}
    \vspace{-10px}
    \label{fig:hoiresult}
\end{figure*}

We evaluate object reconstruction on BEHAVE~\cite{behave} and IMHD$^2$~\cite{imhoi} using object-surface Chamfer distance (CD-o) under two protocols, i.e., Per-frame CD-o (frame-wise holistic Procrustes alignment) and Sliding-window CD-o (holistic Procrustes alignment over combined meshes within a 10-second window). We further compare our method against existing methods. As shown in Table~\ref{tab:cdo_behave_imhd}, 4DHOISolver consistently outperforms all baselines across datasets and evaluation protocols, demonstrating strong cross-dataset robustness. 
We further provide qualitative visualizations on Open4DHOI in Fig.~\ref{fig:hoiresult}.

\begin{table*}[t]
    \centering
    \caption{Quantitative comparison of object tracking accuracy with baseline methods.}
    \label{tab:cdo_behave_imhd}
    \resizebox{0.8\linewidth}{!}{%
    \begin{tabular}{l cccc}
        \toprule
        \multirow{2}{*}{Method} & \multicolumn{2}{c}{BEHAVE} & \multicolumn{2}{c}{IMHD$^2$} \\
        \cmidrule(lr){2-3} \cmidrule(lr){4-5}
        & \makebox[2.8cm][c]{CD-o (per-frame)}
        & \makebox[2.2cm][c]{CD-o (10s)}
        & \makebox[2.8cm][c]{CD-o (per-frame)}
        & \makebox[2.2cm][c]{CD-o (10s)} \\
        \midrule
        PHOSA~\cite{phosa}          & 26.90 & 59.08 & 20.26 & 56.80 \\
        CHORE~\cite{chore}          & 10.02 & 20.32 & 16.81 & 31.76 \\
        VisTracker~\cite{vistracker}     & 8.04  & 8.49  & 23.28 & 18.10 \\
        I'm HOI~\cite{imhoi}        & 7.43  & 4.82  & 6.93  & 8.53  \\
        \midrule
        \textbf{Ours}  & \textbf{4.67} & \textbf{4.42} & \textbf{6.72} & \textbf{5.95}      \\
        \bottomrule
    \end{tabular}%
    }
\end{table*}

\subsection{InterPoint Model Experiments}
\label{sec:interpoint_vlm}
We evaluate our proposed method on our newly constructed Open4DHOI dataset. The dataset is randomly split into three subsets: 360 sequences for training, 45 for validation, and 46 for testing. 

\noindent\textbf{Baseline \& Metrics}.
We compare our method against SOTA interaction contact predictors (LEMON~\cite{lemon}, InteractVLM~\cite{interactvlm}), a human-specific contact model (DECO~\cite{deco}), and an object affordance model (3DAffordance-LLM~\cite{3daffordancellm}). Performance is evaluated across human points, object points, and interaction point pairs. We report Precision, Recall, and F1-score for human points. For object points and point pairs, it is unable to calculate precision, so we only measure Recall. 

\begin{figure}[t]
  \centering
  \begin{minipage}{0.48\textwidth}
    \centering
    \includegraphics[width=\linewidth]{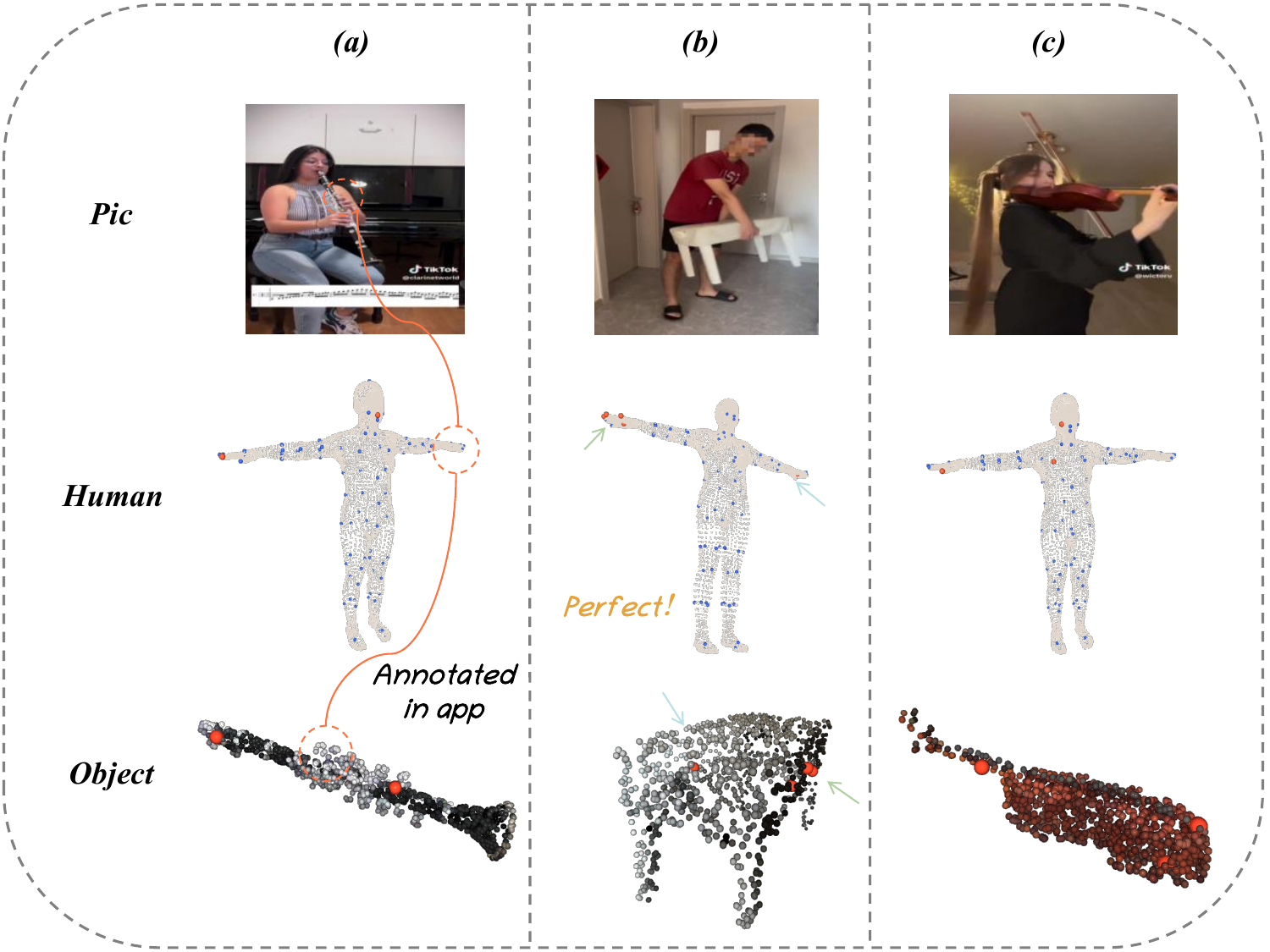}
    \caption{\textbf{InterPoint visualization}: (a): For suboptimal initializations (e.g., missing the left-hand), annotators can refine the annotation in the app. (b),(c): For sufficiently good initializations, annotators can approve them directly. Both cases can greatly speed up the annotation process.}
    \label{fig:network_vis}
  \end{minipage}\hfill 
  \begin{minipage}{0.48\textwidth}
    \centering
    \includegraphics[width=\linewidth]{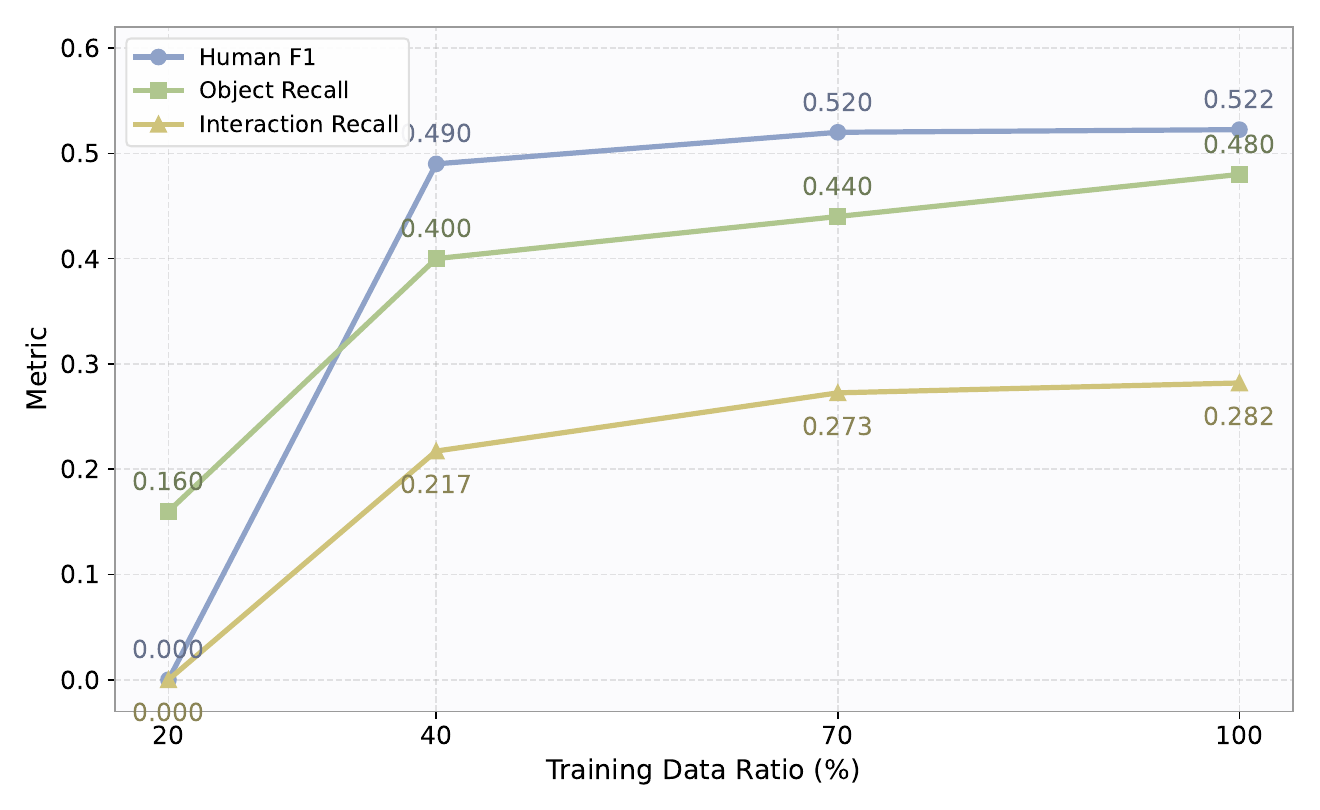}
    \caption{\textbf{Data scalability of Interpoint}: All three metrics increase with more training data, showing that InterPoint improves steadily as the data scale grows. Notably, the simpler human-side task converges faster as more data is added.}
    \label{fig:data_curve}
  \end{minipage}
\end{figure}

The results are shown in Tab.~\ref{tab:point_prediction}. Our model achieves strong performance across all metrics, demonstrating the feasibility of using it for annotation initialization. Qualitative results are presented in Fig.~\ref{fig:network_vis}.

\noindent\textbf{Scalability with Data}.
We aim to use the model to accelerate data generation, while leveraging more data to improve the model and provide better annotation initialization. To this end, we train InterPoint with different data scales (20\%, 40\%, 70\%, and 100\%) and evaluate them using the same metrics, with results presented in Fig.~\ref{fig:data_curve}.
\begin{table*}[t]
    \centering
    \caption{Interaction point prediction results.}
    \label{tab:point_prediction}
    \resizebox{0.75\linewidth}{!}{
    \begin{tabular}{l ccc c c}
        \toprule
        \multirow{2}{*}{Method} & \multicolumn{3}{c}{Human} & \multicolumn{1}{c}{Object} & \multicolumn{1}{c}{Interaction} \\
        \cmidrule(lr){2-4} \cmidrule(lr){5-5} \cmidrule(lr){6-6}
        & Precision $\uparrow$ & Recall $\uparrow$ & F1 $\uparrow$ & Recall $\uparrow$ & Recall $\uparrow$ \\
        \midrule

        DECO~\cite{deco}             & 0.107 & 0.138 & 0.121 & -      & -     \\
        3D-AffordanceLLM~\cite{3daffordancellm} & -     & -     & -     & 0.107  & -     \\
        LEMON~\cite{lemon}            & 0.289 & 0.310 & 0.299 & 0.0894 & 0.045 \\
        InteractVLM~\cite{interactvlm}      & 0.183 & 0.598 & 0.280 & 0.094  & 0.070 \\
        \midrule
        \textbf{Ours}    & \textbf{0.514}  & \textbf{0.532}  & \textbf{0.523}  & \textbf{0.476}   & \textbf{0.303}  \\
        \bottomrule
    \end{tabular}
    }
\end{table*}

\section{Discussion}

While 4DHOISolver is highly scalable, its reliance on temporally invariant interaction points struggles with extreme dynamic sliding contacts (e.g., rolling a walnut in hand) and highly deformable objects (e.g., clothing). Furthermore, our current robotic experiments are validated on a simplified sphere-and-stick humanoid in simulation. Future work will focus on closing the sim-to-real gap by transferring these learned HOI skills to fully actuated, physical humanoids (e.g., Unitree robots), further unlocking our dataset's potential for Embodied AI.

\section{Conclusion}


We present a scalable framework for reconstructing 4D HOI from in-the-wild videos. By leveraging temporally invariant interaction points and an InterPoint-driven human-in-the-loop engine, we efficiently overcome traditional annotation bottlenecks. This enables our 4DHOISolver to perform fast, physically plausible reconstructions. Consequently, we introduce the massive Open4DHOI dataset and demonstrate its significant potential for Embodied AI by training an RL-based humanoid to master complex interactions.

\section*{Supplementary Material}
The contents of this supplementary material are:

Sec.~\ref{sec:interpoint_model}: Details of InterPoint Model

Sec.~\ref{sec:4dhoisolver_supple}: Details of 4DHOISolver.

Sec.~\ref{sec:dataset_suppl}: Characteristics of Open4DHOI.

Sec.~\ref{sec:hoisimulation-supple}: Details of HOI Simulation.

Sec.~\ref{sec:visualizations}: More Visualizations.

\section{Details of InterPoint Model}
\label{sec:interpoint_model}

\subsection{Model Details}
Given an input image \(I\) and an object point cloud \({P}_o\), the model first employs a VLM to extract semantic feature and produces two semantic embeddings:
\begin{equation}
    (E_h, E_o)=\Phi_{\mathrm{VLM}}(I).
\end{equation}
On the geometric side, the object branch passes through an encoder and point feature decoding to obtain object point features and point coordinates. In the human branch, we adopt learnable template parameters defined on the shared SMPL-X topology to generate human point representations, which are then conditioned on the semantic features of the current sample. 

Semantic injection is implemented via a FiLM mechanism rather than simple concatenation. Specifically, the semantic vector is linearly projected to produce channel-wise modulation parameters, which are then applied to the point features:
\begin{equation}
    \hat{\mathbf{f}}=\mathbf{f}\odot (1+\gamma(E))+\beta(E).
\end{equation}
This process is performed separately for the human and object branches, dynamically mapping 2D semantics into the 3D point feature space.

The human point features and object point features are then concatenated into a unified memory and fed into the \textbf{PointInteractionTransformer}. The model contains two groups of learnable queries, corresponding to 87 human keypoints and 87 object contact queries, respectively. Through multiple layers of cross-attention and self-attention, the queries interact with the memory and are iteratively updated, yielding the final human-query and object-query representations.

The prediction stage consists of two heads. The human head outputs a binary logit for each human query, and a sigmoid function is applied to obtain the contact probability for each keypoint. The object head first linearly projects each object query, and then computes scaled dot-product similarities with the object point features, producing classification logits over all object points for each query:
\begin{equation}
    \ell_{k,n}=\frac{(W_q q_k)^\top f_n}{\sqrt{d}},
\end{equation}
where $q_k \in \mathbb{R}^{d}$ denotes the $k$-th object query, $W_q \in \mathbb{R}^{d \times d}$ is the object head, $f_n \in \mathbb{R}^{d}$ is the feature of the $n$-th object point.
Therefore, the object branch is essentially a discrete classification process from each query to the set of point cloud vertices.

During inference, each object query directly selects the point with the maximum response as the predicted contact point index, which is then mapped back to the corresponding point cloud coordinate as the final keypoint prediction:
\begin{equation}
    \hat n_k=\arg\max_n \ell_{k,n}, \qquad \hat x_k=\mathbf{X}_o[\hat n_k].
\end{equation}

Further architectural details of the model are provided in Tab.~\ref{tab:interpoint_config}.

\subsection{Experiment Details}
\label{app:exp_details_metrics}
The baseline predicts dense contact probabilities on vertices for both human and object. After thresholding, we obtain a predicted contact region \(\hat{\mathcal{G}}\), while \(\mathcal{G}\) denotes GT
contact vertices. Its primary metric is point-coverage recall:
\begin{equation}
\mathrm{Recall}_{\mathrm{cov}}=\frac{|\mathcal{G}\cap\hat{\mathcal{G}}|}{|\mathcal{G}|}.
\end{equation}
The other pair-level metrics (micro P/F1) of baseline are computed in similar way, checking whether the GT paired object vertex is covered by
\(\hat{\mathcal{G}}\), with \(\mathrm{PairRecall}_{\mathrm{joint}}\) additionally requiring correct human-contact prediction at the same keypoint.

Our model outputs 87 selected object points (one per query). Let \(\hat{\mathcal{P}}\) be the predicted point set and \(\mathcal{G}\) the GT object-contact points in the same point-cloud
space. We compute distance-based object recall:
\begin{equation}
    \mathrm{Recall}_{\mathrm{obj}}=\frac{1}{|\mathcal{G}|}\sum_{g\in\mathcal{G}}
\mathbf{1}\!\left[\min_{p\in\hat{\mathcal{P}}}\|g-p\|_2<\tau_o\right].
\end{equation}
Our model directly outputs logits for 87 human keypoints as well, so the human metrics are computed in the standard way. Pair metrics are computed under the same distance threshold.
criterion \(\tau_o\).
\begin{table*}[t]
  \centering
  \caption{Configuration of \textbf{InterPoint}.}
  \label{tab:interpoint_config}
  \small
  \begin{tabular}{ll}
  \toprule
  \textbf{Item} & \textbf{Setting} \\
  \midrule
  Backbone VLM & LLaVA-1.5-7B\\
  Object encoder & PointNet++ encoder , PointNet2 feature decoder \\
  PointInteractionTransformer & \(L=6\), \(H=8\), FFN dim \(=1024\), \(d_{\mathrm{tr}}=256\)\\
  Human head & \(256 \rightarrow 128 \rightarrow 1\) \\
  Object head &  \(256 \rightarrow 256\) \\
  Input image size & \(224\times224\) \\
  Object points & \(N_o=1024\) \\
  Batch size & 16 \\
  Learning rate & \(3\times10^{-5}\) \\
  Epochs & 80 \\
  Optimizer / scheduler & AdamW + cosine annealing \\
  Gradient clipping & Global norm \(=1.0\) \\
  Contrastive temperature & \(\tau_c=0.1\) \\
  Loss weights & \(\lambda_h=1.0,\ \lambda_{k2o}=1.0,\ \lambda_{o2k}=0.3\) \\
  \bottomrule
  \end{tabular}
\end{table*}
\section{Details of 4DHOISolver}
\label{sec:4dhoisolver_supple}

\subsection{Human Keypoint Defination}

In Sec.~3.2, we mention that, for convenient and accurate interaction annotation, we predefine keypoints on the human body to serve as annotation targets. We first subdivide the human body parts by evenly splitting each major joint into front, back, left, and right regions. Then, for each subdivided part, we select a central point as the interaction keypoint. For the hand interaction points, we selected corresponding locations on all five fingers, as well as points on both the palm and the back of the hand.

\begin{figure}[t]
    \centering
    \begin{minipage}[c]{0.52\linewidth}
        \centering
        \small
        \makeatletter\def\@captype{table}\makeatother
        \caption{Human joint-tree.}
        \label{tab:joint_tree_clean}
        \resizebox{\linewidth}{!}{
            \begin{tabular}{l c}
            \toprule
            \textbf{Main-Joints} & \textbf{Sub-Joints} \\
            \midrule
            leftForeArm & back, pinky, wrist, thumb \\
            rightForeArm & back, pinky, wrist, thumb \\
            leftUpperArm & up, down, back, front \\
            rightUpperArm & up, down, back, front \\
            leftShoulder & front, back \\
            rightShoulder & front, back \\
            leftHand & back, palm, Thumb, Index, Middle, Ring, Pinky \\
            rightHand & back, palm, Thumb, Index, Middle, Ring, Pinky \\
            leftUpperLeg & inner, outer, front, back \\
            rightUpperLeg & inner, outer, front, back \\
            leftLowerLeg & front, outer, back, inner \\
            rightLowerLeg & front, outer, back, inner \\
            leftFoot & ToeBase, instep, sole \\
            rightFoot & ToeBase, instep, sole \\
            upperSpine & back, right, front, left \\
            middleSpine & front, right, back, left \\
            leftNeck & front, back \\
            rightNeck & back, front \\
            hip & front, left, front, back \\
            buttocks\_left & buttocks\_left \\
            buttocks\_right & buttocks\_right \\
            head & mouth, chin, headtop, cheek\_left, cheek\_right \\
            rightElbow & back, front \\
            leftElbow & back, front \\
            leftKnee & back, front \\
            rightKnee & back, front \\
            \bottomrule
            \end{tabular}
        }
    \end{minipage}
    \hfill
    \begin{minipage}[c]{0.44\linewidth}
        \centering
        \includegraphics[width=\linewidth]{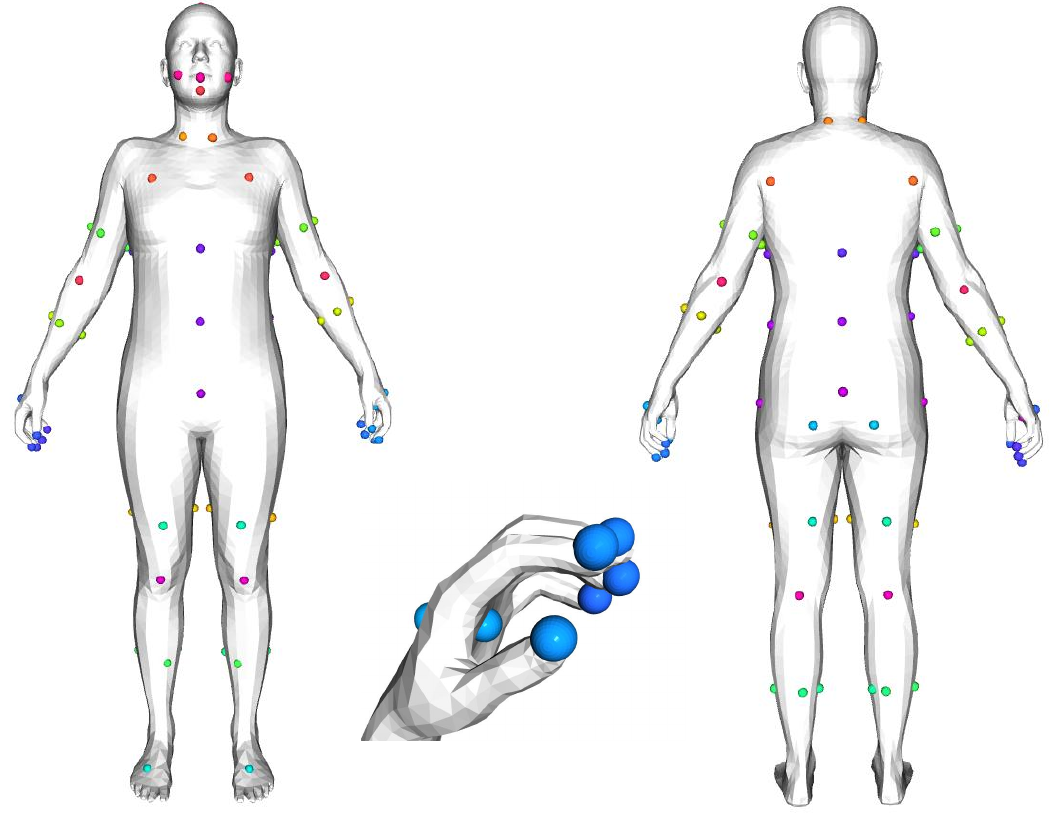}
        \caption{Human keypoint definition.}
        \label{fig:human_keypoint}
    \end{minipage}
\end{figure}

In total, we defined 87 keypoints. Fig.~\ref{fig:human_keypoint} visualizes our human body keypoints, and Tab.~\ref{tab:joint_tree_clean} lists the names of the joints.

\subsection{Preprocess App}
To streamline the collection of in-the-wild videos, we develop a custom video processing application. Specifically, we first download candidate videos from platforms such as TikTok and record their source URLs. To ensure dataset diversity and prevent redundancy, each URL is queried against our database for deduplication; only newly discovered videos are retained and saved. Annotators then assign specific object categories to each video. Since internet videos frequently contain rapid scene changes, we apply an automatic shot transition detection algorithm to temporally segment the raw video into continuous, single-shot clips. From these, annotators select up to three high-quality clips per video that best capture the target human-object interactions. 

\begin{figure}[t]
    \centering
    \begin{minipage}[c]{0.48\linewidth}
        \centering
        \includegraphics[width=\linewidth]{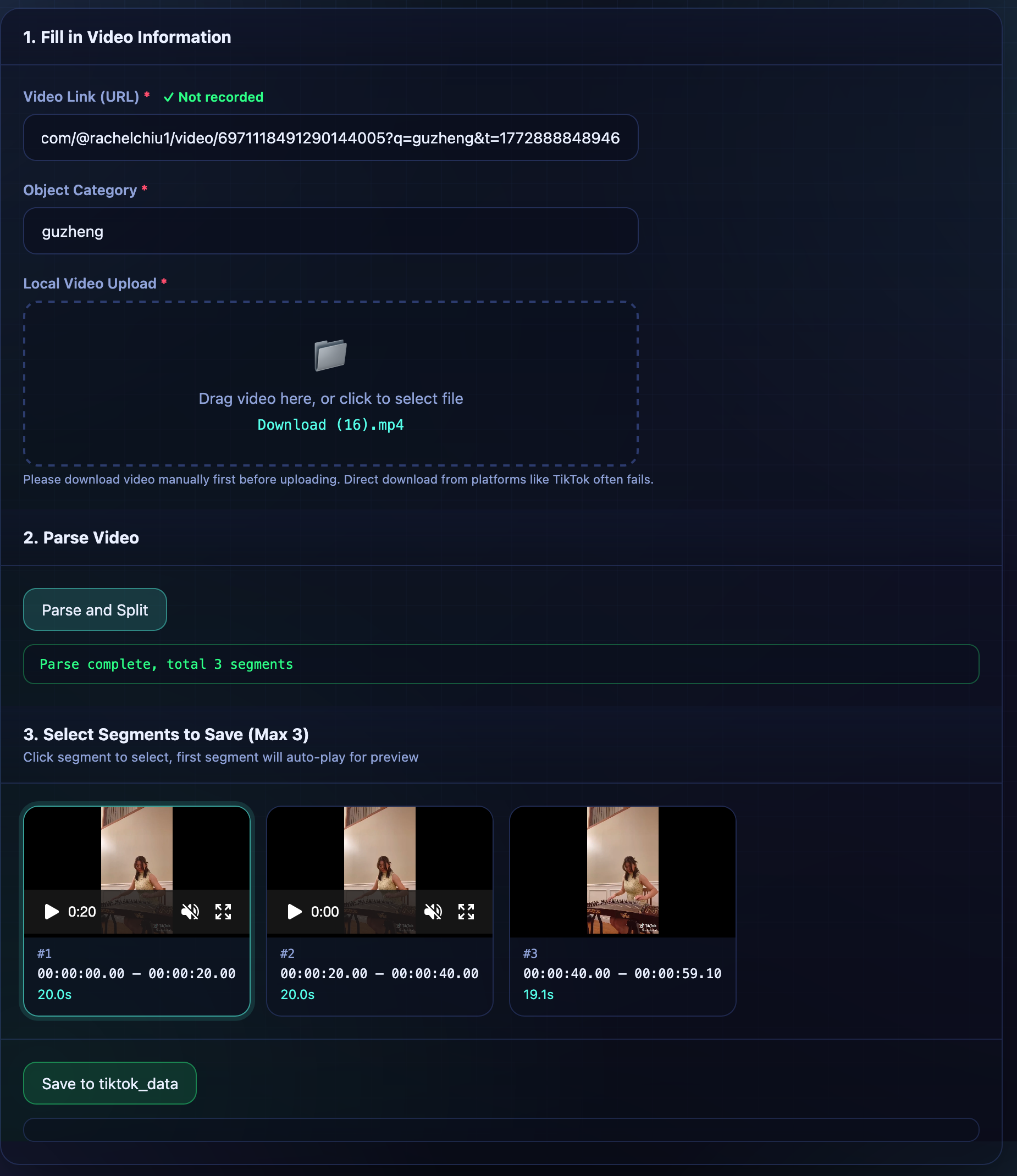} 
        \caption{User interface of our video collection and processing application.}
        \label{fig:upload_app}
    \end{minipage}
    \hfill
    \begin{minipage}[c]{0.48\linewidth}
        \centering
        \includegraphics[width=\linewidth]{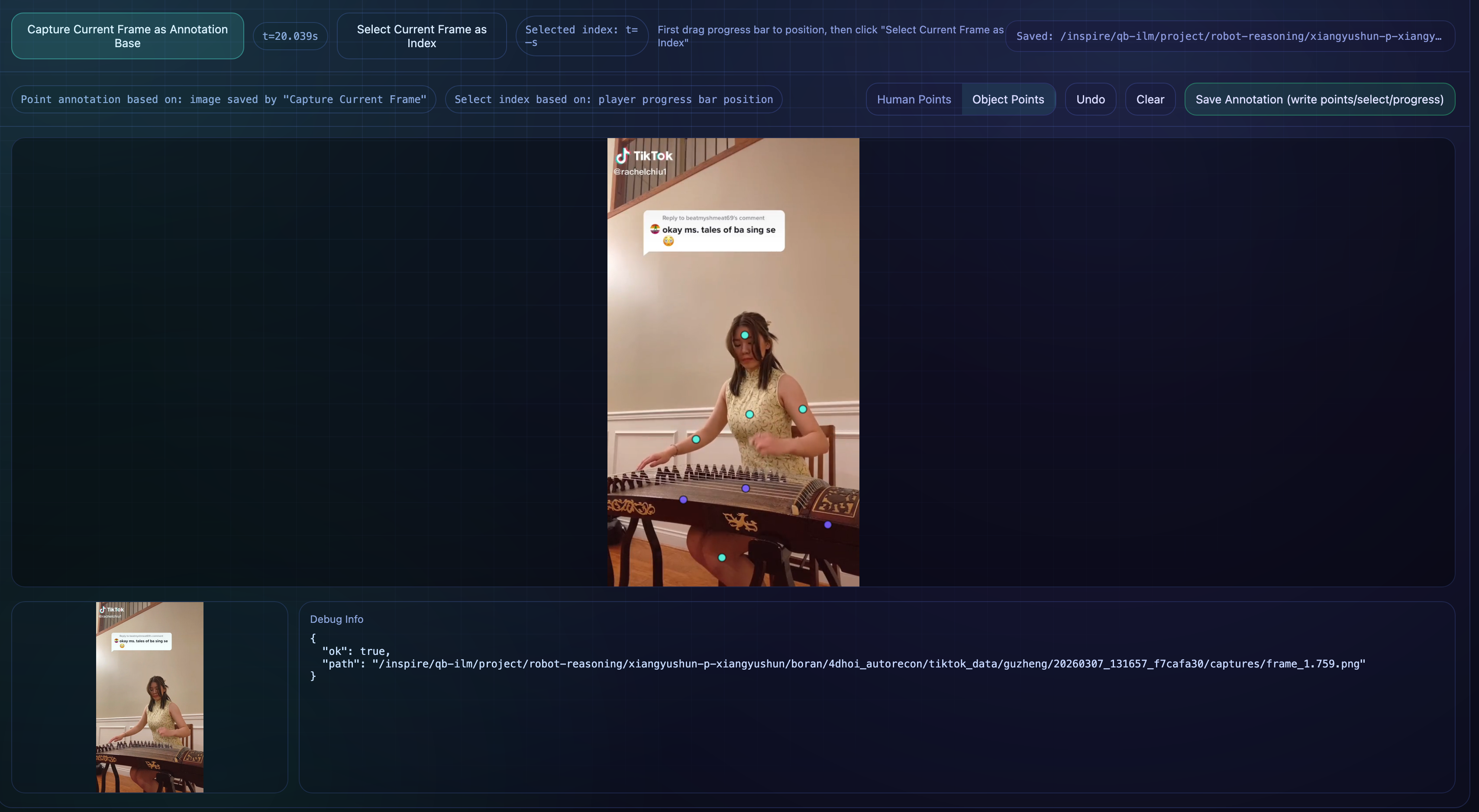} 
        \caption{Our pre-process workflow displaying prompt points and SAM2 segmentation.}
        \label{fig:preprocess_app}
    \end{minipage}
\end{figure}

Following this data curation pipeline, as mentioned in Sec.~3.1, video tracking is required to obtain spatio-temporally consistent mask sequences for the subsequent 4D reconstruction. For each selected clip, we identify the starting frame of the interaction and manually provide point prompts for both the human and the object. These prompts are then fed into SAM2~\cite{sam2} to extract the complete mask sequences. As illustrated in Fig.~\ref{fig:preprocess_app}, our annotation workflow displays the segmentation results of the separately annotated prompt points to guarantee accurate tracking before proceeding to the 4DHOISolver.

\subsection{Annotation App}
\label{supple:app}
\begin{figure}[!t]
    \centering
    \includegraphics[width=0.9\linewidth]{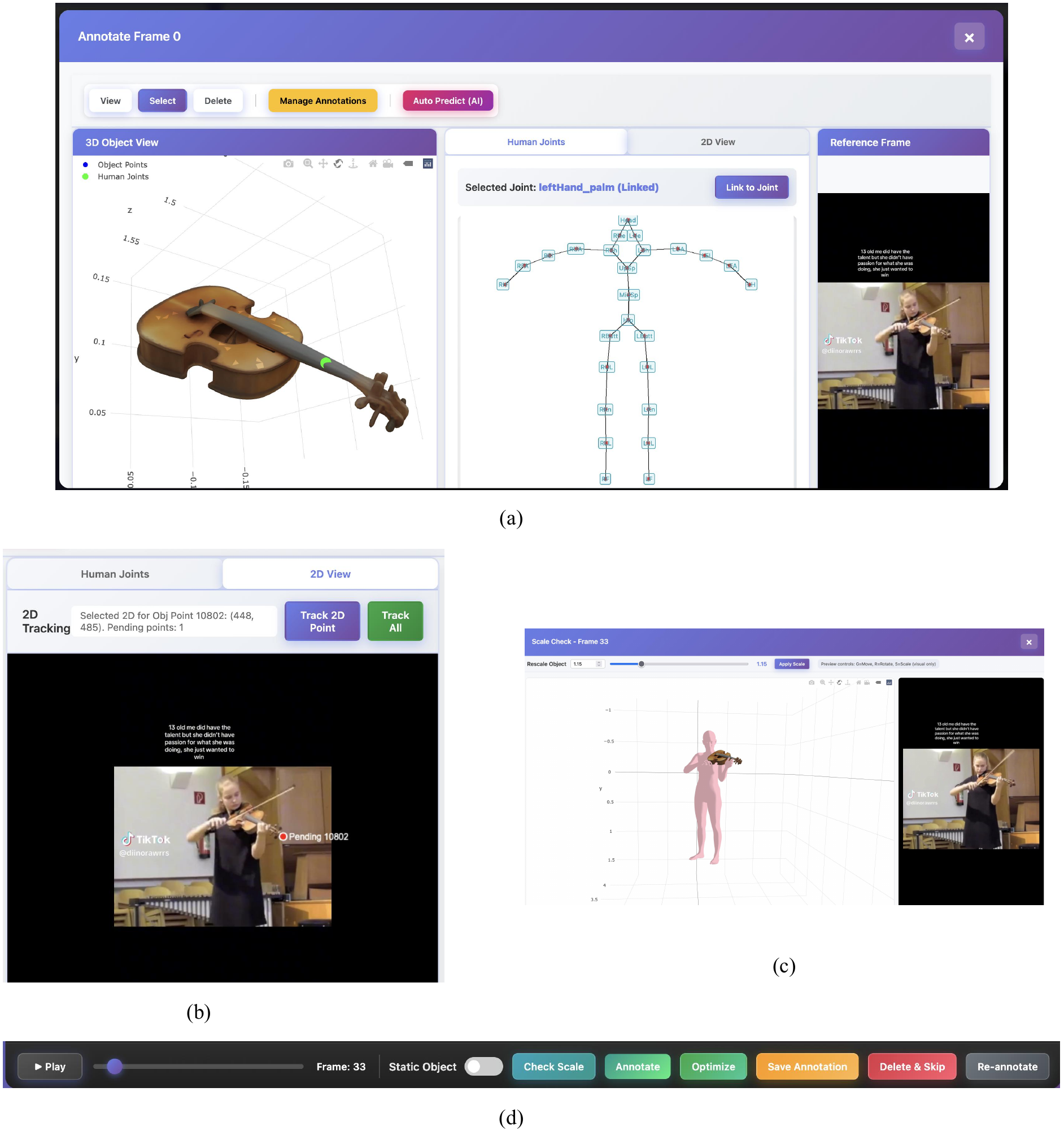}
    \caption{\textbf{Overview of our interactive annotation system.} (a) The interface for selecting object interaction points and assigning corresponding human joints. (b) The 2D point tracking interface for verifying temporal correspondences. (c) The viewer for inspecting the fast optimization and 4D reconstruction results. (d) The interactive toolbar for system controls.}
    \label{fig:supple_app}
\end{figure}

\begin{figure}[h]
    \centering
    \includegraphics[width=\linewidth]{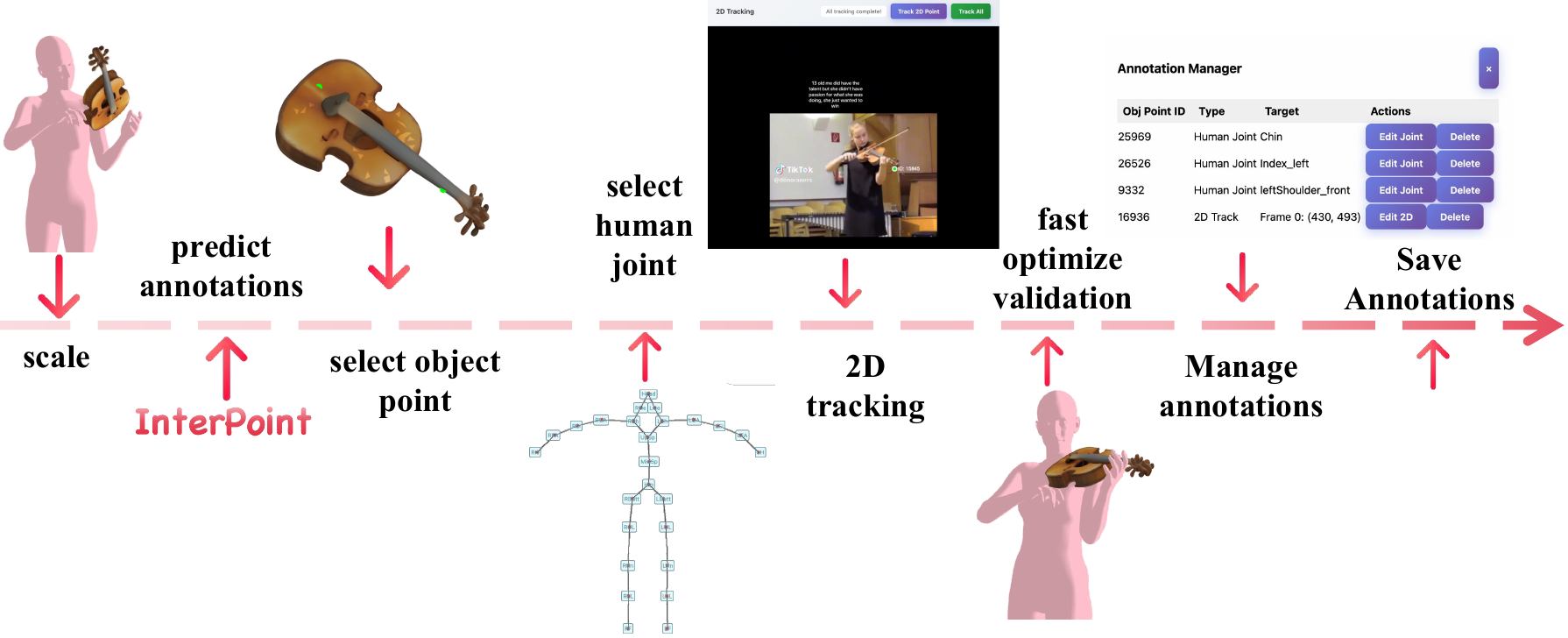}
    \caption{\textbf{Workflow of our human-in-the-loop annotation pipeline.} The timeline illustrates the step-by-step process from initial video preprocessing and interactive point annotation to the final 4D HOI reconstruction.}
    \label{fig:supple_app_workflow}
\end{figure}

To efficiently collect high-quality 4D HOI data, we develop a custom interactive annotation application. Fig.~\ref{fig:supple_app} displays the user interface, and Fig.~\ref{fig:supple_app_workflow} illustrates the step-by-step workflow.

\vspace{1mm}\noindent\textbf{User Interface \& Toolbar.} Our system provides an intuitive toolbar for rapid interaction control:
\begin{itemize}
    \setlength{\itemsep}{1pt}
    \setlength{\parskip}{0pt}
    \setlength{\parsep}{0pt}
    \item \textbf{Video Timeline:} Users can drag the progress bar or play the video to locate specific frames that require annotation.
    \item \textbf{Check Scale:} This function allows annotators to inspect and adjust the automatically reconstructed object scale, as well as visually verify the optimization results.
    \item \textbf{Static Object:} A toggle designed for stationary items. When activated, the system fixes the object's pose for the entire video using the annotation from the frame with the maximum number of labeled points, significantly reducing redundant manual effort.
    \item \textbf{Optimize:} Triggers our HOI keypoint solver for rapid geometric alignment. This process takes only about 5 seconds on average, providing immediate visual feedback on the annotation quality.
\end{itemize}

\vspace{1mm}\noindent\textbf{Annotation Workflow.} As shown in the workflow diagram, a complete annotation cycle proceeds as follows:
\begin{itemize}
    \setlength{\itemsep}{1pt}
    \setlength{\parskip}{0pt}
    \setlength{\parsep}{0pt}
    \item \textbf{InterPoint Auto-prediction:} At any given frame, annotators can invoke our InterPoint model to automatically predict initial human-object contact point pairs, drastically reducing from-scratch manual selection.
    \item \textbf{3D Point \& Human Joint Selection:} Based on the AI proposals, users can refine the results by selecting 3D contact points on the object and assigning them to corresponding human joints via a hierarchical joint-tree menu. 
    \item \textbf{2D Point Tracking:} Users specify the 2D image projections of the selected 3D object points. The app then tracks these points temporally, allowing annotators to easily correct any tracking drift.
    \item \textbf{Manage Annotations:} Annotators can flexibly edit, re-link, or delete specific contact pairs or tracking points without discarding the valid parts, greatly improving efficiency.
    \item \textbf{Fast Optimization \& Verification:} Finally, users utilize the \textit{Optimize} function to quickly solve the 4D HOI reconstruction. This allows them to seamlessly verify the physical plausibility and temporal consistency of their current annotations and make further adjustments if necessary.
\end{itemize}

\subsection{Optimization}

\subsubsection{Loss}
In the HOI Optimizer, we use mask loss, contact loss, and collision loss for optimization. Here, we provide a detailed explanation of how these loss functions are employed to enforce the physical plausibility of interactions.

\noindent \textbf{Mask Loss.}
Let $\hat{M}_h$, $\hat{M}_o$, $M_h$, and $M_o$ denote the rendered and ground-truth masks for the human and object, respectively.
To handle mutual occlusions, we compute
\begin{equation}
\tilde{M}_h = \hat{M}_h (1 - M_o), \qquad
\tilde{M}_o = \hat{M}_o (1 - M_h).
\end{equation}

We supervise the silhouettes with a pixel-wise MSE:
\begin{equation}
\mathcal{L}_{\text{mask}} =
\mathrm{MSE}(\tilde{M}_h, M_h)
+
\mathrm{MSE}(\tilde{M}_o, M_o).
\end{equation}

To enhance boundary accuracy, we extract edges via
\begin{equation}
E_h = \mathrm{Pool}(\tilde{M}_h) - \tilde{M}_h, \qquad
E_o = \mathrm{Pool}(\tilde{M}_o) - \tilde{M}_o,
\end{equation}
and compute distance-transform weights $W_h$, $W_o$ on the 
ground-truth edges.  
The edge loss is
\begin{equation}
\mathcal{L}_{\text{edge}}
= 
\sum E_h W_h
+
\sum E_o W_o .
\end{equation}

The final loss is
\begin{equation}
\mathcal{L}_{\text{total}}
=
\alpha \mathcal{L}_{\text{mask}}
+
\beta \mathcal{L}_{\text{edge}},
\end{equation}
where $\alpha$ and $\beta$ are non-negative weights that balance the contributions of the mask and edge terms.

\noindent \textbf{Contact Loss.}
Given corresponding human-object point pairs $\{(p_i^{b}, p_i^{o})\}_{i=1}^{N}$, we compute their Euclidean distances for each pair as $d_i = \| p_i^{b} - p_i^{o} \|_2$.
To encourage all pairs to converge to a globally balanced contact configuration,
we assign larger weights to pairs with larger distances:
\begin{equation}
w_i = \frac{(d_i + \epsilon)^2}{\sum_{j=1}^{N} (d_j + \epsilon)^2}.
\end{equation}
This weighting scheme pulls distant pairs more strongly while preventing near-contact pairs from dominating the gradients, leading to a uniform and stable convergence of all contact points.

The final contact loss is defined as:
\begin{equation}
\mathcal{L}_{\text{contact}}
=
\sum_{i=1}^{N} 
w_i \, d_i^{\,2}.
\end{equation}

\noindent \textbf{Collision Loss.}
To prevent interpenetration between the human mesh and the object mesh, 
we adopt the bidirectional mesh-to-mesh collision penalty used in~\cite{collision}. 
Specifically, we apply the same collision operator to measure 
(i) human vertices inside the object surface and 
(ii) object vertices inside the human surface. 
The final collision loss is a weighted combination
\begin{equation}
    \mathcal{L}_{\text{coll}} = 
\lambda_{\text{h}\rightarrow\text{o}} \,\mathcal{L}_{\text{h}\rightarrow\text{o}}
\;+\;
\mathcal{L}_{\text{o}\rightarrow\text{h}},
\end{equation}
where $\lambda_{\text{h}\rightarrow\text{o}}$ controls the relative importance of penalizing
human–inside–object penetration.

\subsubsection{Static Strategy}
For videos labeled with the static-object option, we identify the frame with the largest number of annotated interaction points and use it as the static optimization frame. The object pose optimized in this frame is then fixed, and the object no longer participates in subsequent optimization steps. In the subsequent optimization process, only the human parameters are optimized.

\subsection{More Visualizations}
\label{supple:more_vis}

\begin{figure}[H]
    \centering
    \includegraphics[width=0.9\linewidth]{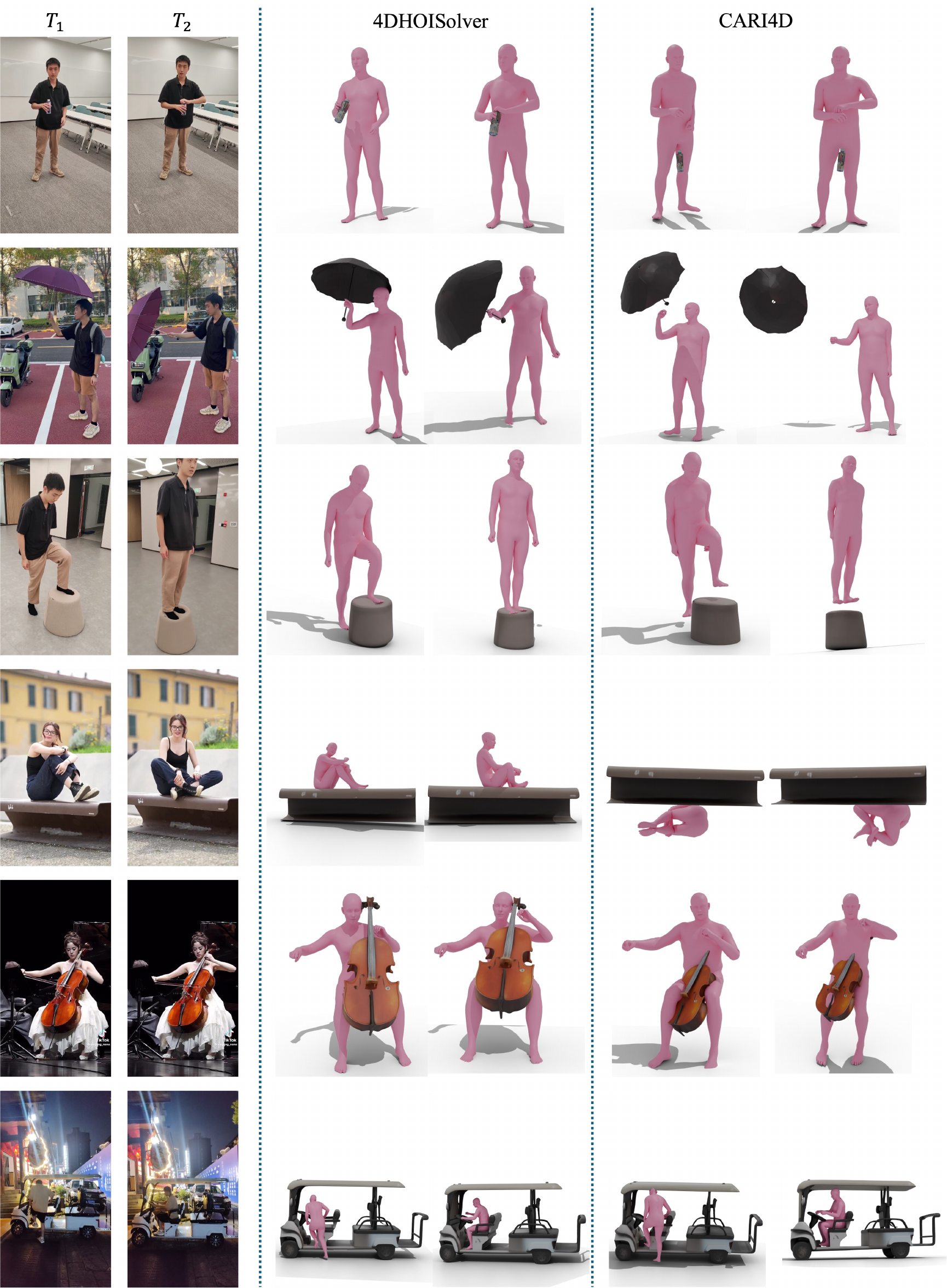}
    \caption{More Visualizations for 4DHOISolver Reconstructions.}
    \label{fig:supple_vis_recon}
\end{figure}

Fig.~\ref{fig:supple_vis_recon} showcases additional 4D reconstruction results of our 4DHOISolver on diverse in-the-wild videos, together with comparisons against CARI4D~\cite{cari4d}. For object reconstruction, we use the object meshes obtained by our SAM 3D Objects~\cite{sam3dobjects} pipeline, i.e., the same object meshes as those used in 4DHOISolver, to eliminate the influence of object geometry differences. For human reconstruction, we directly follow the original CARI4D pipeline, including UniDepth~\cite{unidepth} for monocular depth estimation, Neural Localizer Fields (NLF)~\cite{nlf} for human pose prediction, SMPL-H global parameter fitting, FoundationPose~\cite{foundationpose} for object tracking, and CoCoNet~\cite{cari4d} for joint optimization. We use the officially released model weights and default settings throughout. To ensure a fair comparison, both methods take the same input videos and identical object meshes as input. As illustrated, our method produces better human--object spatial alignment and more physically plausible contacts, effectively reducing common artifacts such as object floating and severe interpenetration. In addition, our reconstructions exhibit stronger temporal consistency and more faithful interaction geometry across frames. Overall, these examples highlight the advantage of 4DHOISolver in recovering realistic, stable, and coherent 4D human--object interactions in unconstrained real-world videos.

\subsection{Details for 4DHOISolver Experiments}
On both BEHAVE~\cite{behave} and IMHD$^2$~\cite{imhoi} datasets, since 4DHOISolver jointly relies on 3D--3D human--object contact correspondences and 3D--2D object observation constraints, we first extract sparse constraints from the raw annotations in a manner consistent with our formulation. For the 3D--3D contact points, within the human--object interaction region, we use each SMPL body part as a query and search for its nearest point on the object surface. If the Euclidean distance is smaller than a predefined threshold ($0.1$ m), the pair is treated as a valid body--object contact correspondence. We then sort all candidate pairs by distance and keep at most 5 of them. In this way, only sufficiently close pairs are regarded as effective contacts, providing strong constraints on the relative human--object pose while reducing sensitivity to noise and mismatches through sparsification. For the 3D--2D object points, we apply farthest point sampling on the projected object mesh and select 5 3D--2D correspondences that are as uniformly distributed as possible in the image plane. These sparse yet informative 2D observations provide additional constraints on the object position, scale, and depth. Overall, this sparse constraint design improves robustness to noise and occlusion and stabilizes the optimization, while still providing sufficient geometric guidance for reconstruction.

We evaluate object reconstruction using object Chamfer Distance (CD-o). For each frame, we uniformly sample $2\times10^4$ points from each of the ground-truth and predicted object meshes, denoted by $P_{gt}$ and $P_{pred}$, respectively, and define
\begin{equation}
    \mathrm{CD\mbox{-}o}
=
\frac{1}{|P_{pred}|}\sum_{x\in P_{pred}}\min_{y\in P_{gt}}\|x-y\|
+
\frac{1}{|P_{gt}|}\sum_{y\in P_{gt}}\min_{x\in P_{pred}}\|y-x\|,
\end{equation}
where the CD-o metric is evaluated following the protocol used in VisTracker~\cite{vistracker}. We then average the distance over all evaluation frames to obtain the final object reconstruction error.

\section{Characteristics of Open4DHOI}
\label{sec:dataset_suppl}

\subsection{Data Collection}


To obtain high-quality HOI motions, the data should satisfy several requirements. First, the input videos should be clear and high-resolution, as reliable visual details are critical for both object and human reconstruction. 

Second, the data should cover diverse scenes and a wide range of interaction types in order to capture rich and representative HOI patterns. 

Third, we focus on full-body third-person-view interaction data, since existing human reconstruction methods still have difficulty accurately recovering certain body poses under partial visibility and self-occlusion.

We recruited volunteers to collect short video data from TikTok, requiring the clips to satisfy exocentric viewpoints and full-body interactions. In addition, we adopted a two-person cooperative recording setup, where one volunteer captured the scene with a mobile phone while the other performed interactions with a designated object. These two methods are shown in Fig.~\ref{fig:data_collection}.



\subsection{Action Annotation}
In Sec.~5.1.1, we mention that we annotated 133 action categories for our dataset. We automatically extracted these action categories using a two-step strategy involving Qwen2.5-VL-72B~\cite{qwen2.5vl} and Qwen2.5-72B-Instruct~\cite{qwen2.5}, followed by manual filtering.

First, we used Qwen2.5-VL-72B to extract interaction descriptions from the videos. The prompt we provided was:
``There exists human and \{obj\_name\} in the video. 
Please describe the interactions between the person and the object 
completely and accurately. Output full sentences.''

Next, we used Qwen2.5-72B-Instruct to automatically extract the action categories. The prompt we provided was: 
``Here is a human-\{obj\_name\} action description: \{interaction\_description\}. Please summarize the core human-\{obj\_name\} interaction into several concise action category names.''

After the above two-step extraction process, we further performed manual filtering and consolidation on the generated action names to remove redundant, overly specific, or semantically ambiguous categories. As a result, we obtained a structured action vocabulary for our dataset. This vocabulary provides the basis for subsequent action annotation and analysis, while the overall semi-automatic pipeline significantly reduces annotation cost and improves labeling efficiency without sacrificing the diversity of human--object interactions.


\subsection{Co-occurence of HOI}
\begin{figure}[t]
    \centering
    \begin{minipage}[b]{0.48\linewidth}
        \centering
        \includegraphics[width=\linewidth]{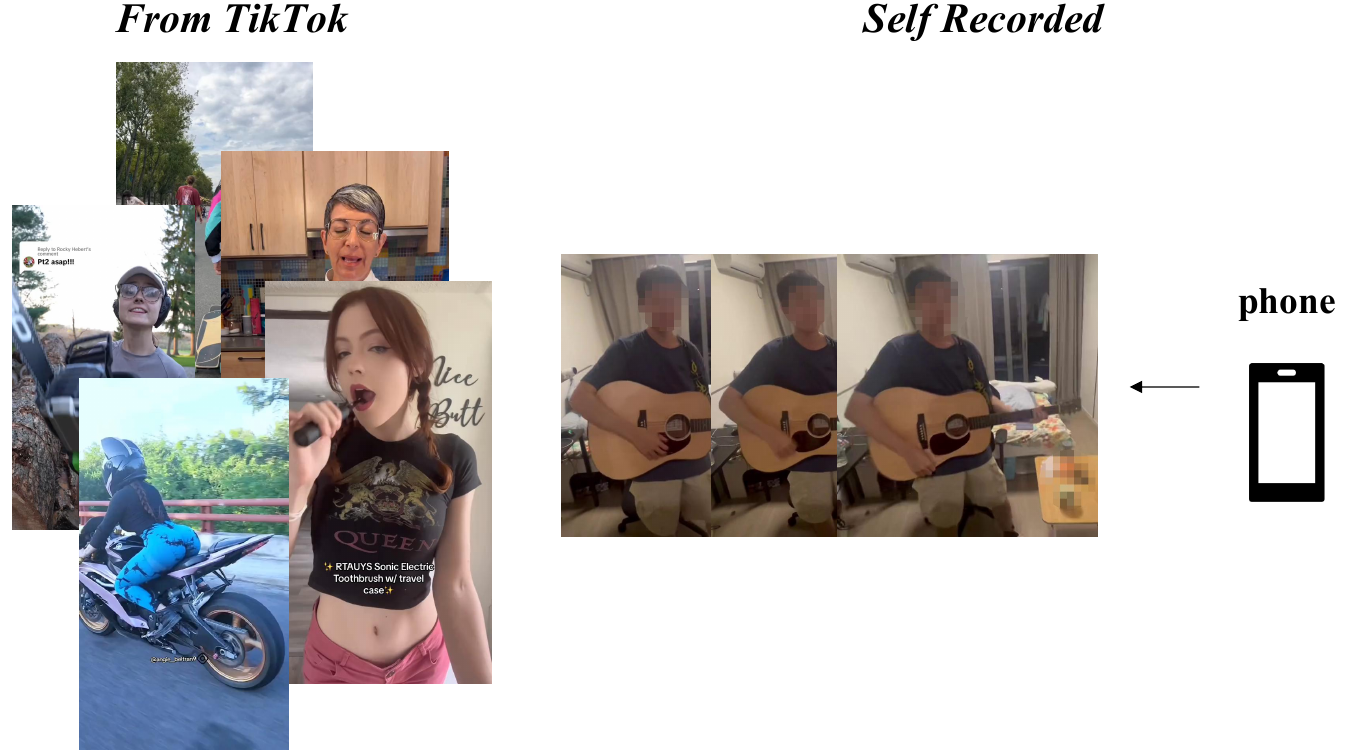}
        \caption{Data collection methods.}
        \label{fig:data_collection}
    \end{minipage}
    \hfill
    \begin{minipage}[b]{0.48\linewidth}
        \centering
        \includegraphics[width=\linewidth]{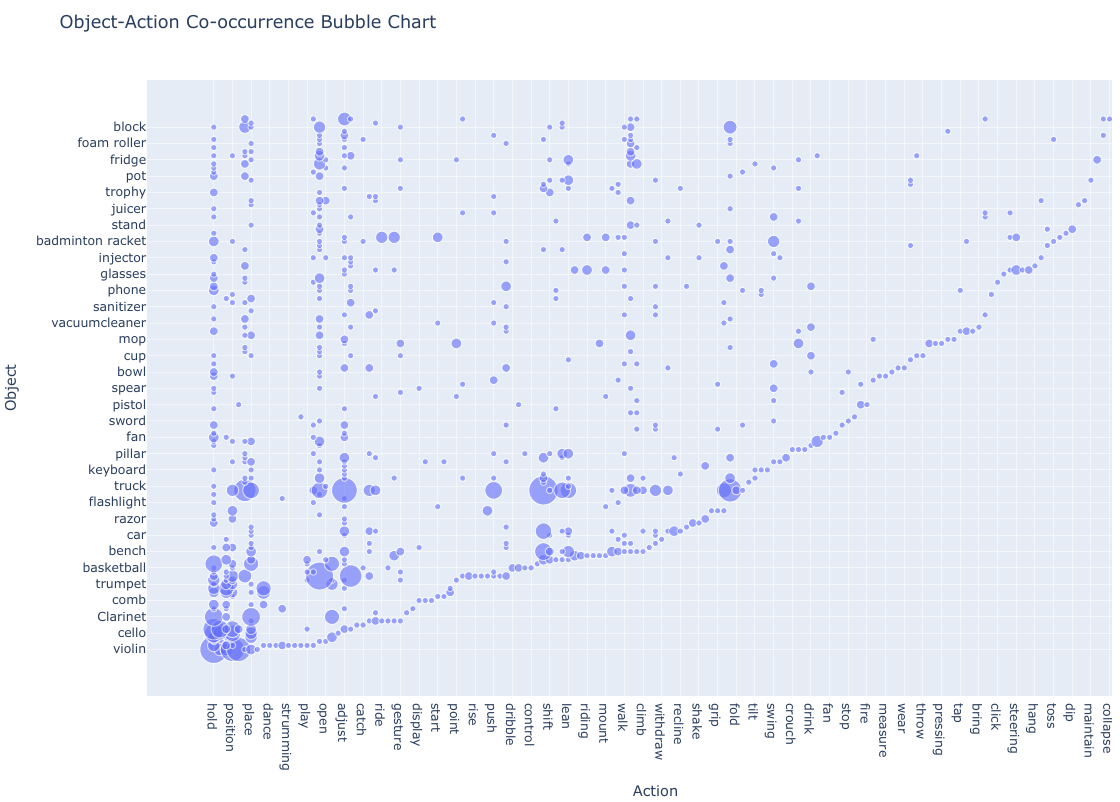}
        \caption{Co-occurrence between object categories and actions in Open4DHOI.}
        \label{fig:supple_coocurrence}
    \end{minipage}
\end{figure}

We plotted the co-occurrence matrix between actions and object categories, as shown in Fig.~\ref{fig:supple_coocurrence}. It can be observed that our dataset contains many hand-related interaction objects, including actions such as ``pick up'' and ``hold'', as well as frequent interactions like ``sit'' and ``stand on''. This also demonstrates that our dataset covers a highly diverse range of HOI categories.

\section{Details of HOI Simulation}
\label{sec:hoisimulation-supple}
\subsection{Implementation Details}
The reference data is derived from human bodies represented using SMPL-X~\cite{smpl-x}. For simulation, we retarget these models into rigid bodies following~\cite{Luo2023PerpetualHC,yuan2021simpoesimulatedcharactercontrol}, and Objects are also converted into simulation models through convex decomposition.
Similar to existing methods~\cite{wang2023physhoiphysicsbasedimitationdynamic,xu2025intermimic}, we perform HOI-simulation in Isaac Gym and use the first reference frame to initialize the simulation environment. The parameter settings for simulation environments are shown in the Tab.~\ref{tab:Issacgym_hyperparameter} To distinguish between static and dynamic objects derived from the annotation 
\begin{table}[!t]
    \centering
    \caption{Simulation hyperparameters.}
    \resizebox{0.45\linewidth}{!}{%
    \begin{tabular}{lc}
        \toprule
        \textbf{Hyperparameter} & \textbf{Value} \\
        \midrule
        Sim dt                       & 1/60s \\
        Control dt                   & 1/30s \\
        Number of envs               & 1024 \\
        Number of substeps           & 4 \\
        Number of pos iterations     & 8 \\
        Number of vel iterations     & 1 \\
        Contact offset               & 0.2 \\
        Rest offset                  & 0.0 \\
        Max depenetration velocity   & 20 \\
        Object friction              & 0.6 \\
        Object static mass           & 10000 \\
        Object dynamic mass          & 0.5 \\
        Object \& ground restitution & 0.05 / 0.1 \\
        Object density               & 1000 \\
        Object max convex hulls      & 64 \\
        \bottomrule
    \end{tabular}%
    }
    \label{tab:Issacgym_hyperparameter}
\end{table}
information, we set different physical masses for them. This ensures physical plausibility while satisfying the static object constraint as much as possible.

\subsection{Metrics}
As described in the Sec.~4, we divided 80 sequences into 12 subsets by action type and trained one policy per subset. To validate our method, we tested and computed three metrics: MPJPE, contact score, and jitter. In this section, we provide further details about how these metrics are calculated.
MPJPE (Mean Per Joint Position Error) is used to measure the average distance between the skeleton joint positions in simulation and the reference joint positions. It directly reflects the overall accuracy of pose or trajectory reconstruction. A smaller value indicates a more accurate reconstruction.
\begin{align}
    \mathrm{MPJPE}=\frac{1}{T|\mathcal{J}|}\sum_{t=1}^{T}\sum_{j=1}^{\mathcal{J}}\left\|\hat{p}^h_{t,j}-p^h_{t,j}\right\|_2,
\end{align}
where $T$ denotes the total number of frames, $\mathcal{J}$ represents the set of human joints, and $|\mathcal{J}|$ indicates its cardinality. The term $\hat{p}^h_{t,j}$ refers to the simulated 3D position of joint $j$ at frame $t$, while $p^h_{t,j}$ denotes the corresponding ground-truth joint position.


The contact score is computed as the sum of squared Euclidean distances between each pair of annotated keypoints in the interaction graph $I_g$. During retargeting, the SMPL-X keypoints that we annotated are converted into positions of skeleton joints. This score is used to check whether the newly added reward successfully improves the contact relationships during simulation.
\begin{align}
    \mathrm{Contact}=\sum \left\| \mathbf{\theta}_t^o\,M^o(I_g) + {p}_t^o - {p}_t^h(I_g) \right\|^2,
\end{align}
where $\mathbf{\theta}_t^o$ and $p_t^o$ represent the object’s rotation and translation at frame $t$, $M^o(I_g)$ selects the object keypoints defined by the interaction graph $I_g$, and $p_t^h(I_g)$ denotes the corresponding human joint positions.

The jitter score is used to measure the smoothness and stability of a sequence over time, and indicates whether noticeable jitter is present. A lower value indicates smoother, more coherent motion. The calculation is as follows:
\begin{align}
    \mathrm{Jitter}=\frac{1}{|\mathcal{J}|}\sum_{j \in \mathcal{J}}\sqrt{\frac{1}{T-3}\sum_{t}\left\|\Delta^{3}\mathbf{p}_{t,j}^h\right\|^{2}}.
\end{align}

\section{More Visualizations}
\label{sec:visualizations}



\subsection{InterPoint Visualizations}
\begin{figure}[H]
    \centering
    \includegraphics[width=0.9\linewidth]{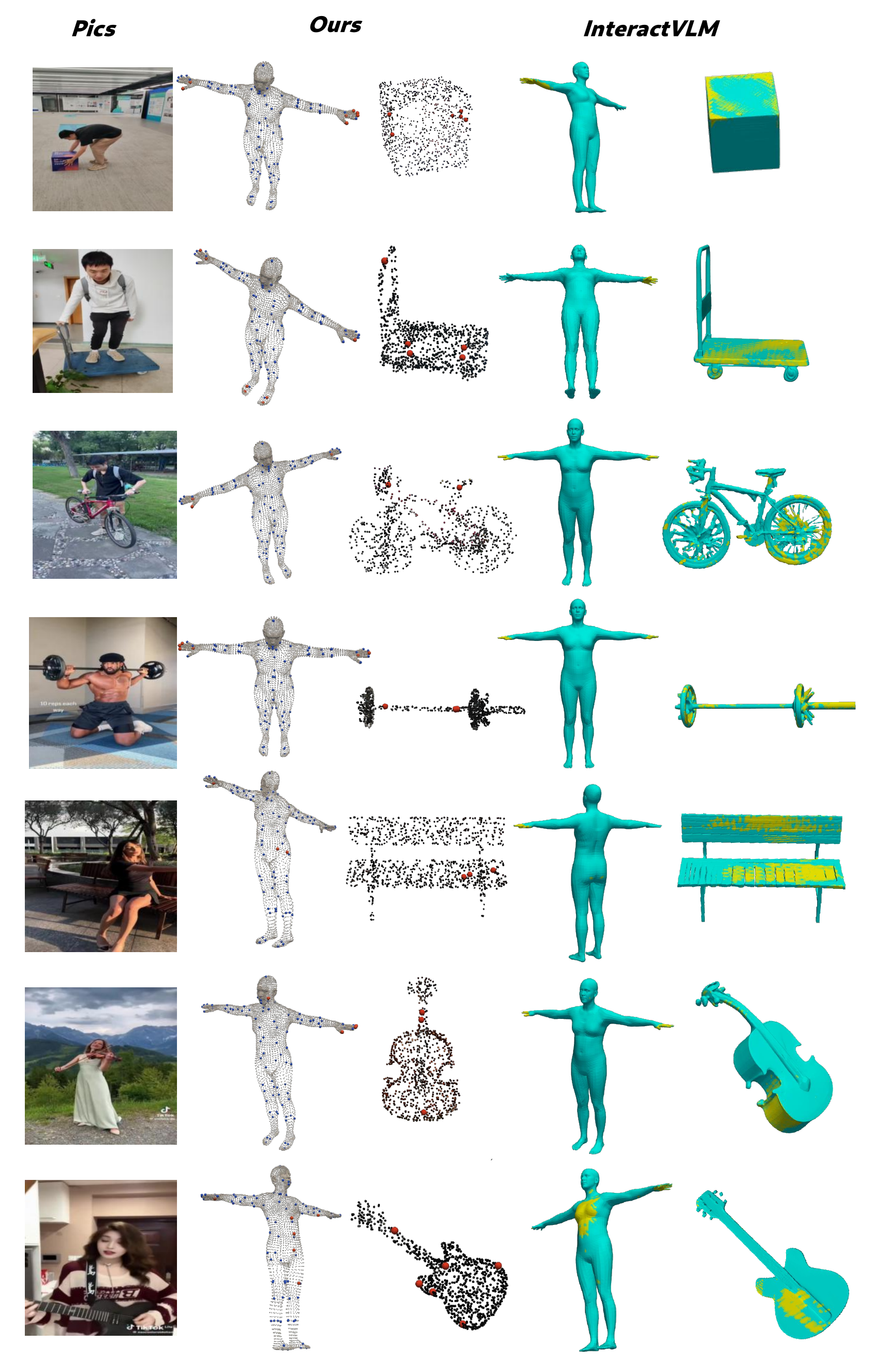}
    \caption{More Visualizations for Interpoint and InteractVLM}
    \label{fig:supple_vis}
\end{figure}

We present more visualizations of InterPoint in Fig.~\ref{fig:supple_vis}, together with the corresponding visualizations of InteractVLM~\cite{interactvlm} on the same samples for comparison. As shown in the figure, our method is able to provide relatively accurate annotation initialization in most cases, especially for identifying plausible human--object contact regions and establishing meaningful point-level correspondences between the human body and object surface. Compared with InteractVLM, InterPoint generally produces predictions that are more spatially consistent with the underlying interaction geometry and better aligned with the actual contact patterns. The predicted correspondences are also more concentrated around semantically and physically reasonable interaction areas, which makes the initialization more suitable for downstream processing. These results further demonstrate that InterPoint can serve as a reliable initialization module, providing high-quality starting annotations for subsequent refinement and optimization.

\subsection{HOI Simulation Visualizations}
\begin{figure}[H]
    \centering
    \includegraphics[width=0.9\linewidth]{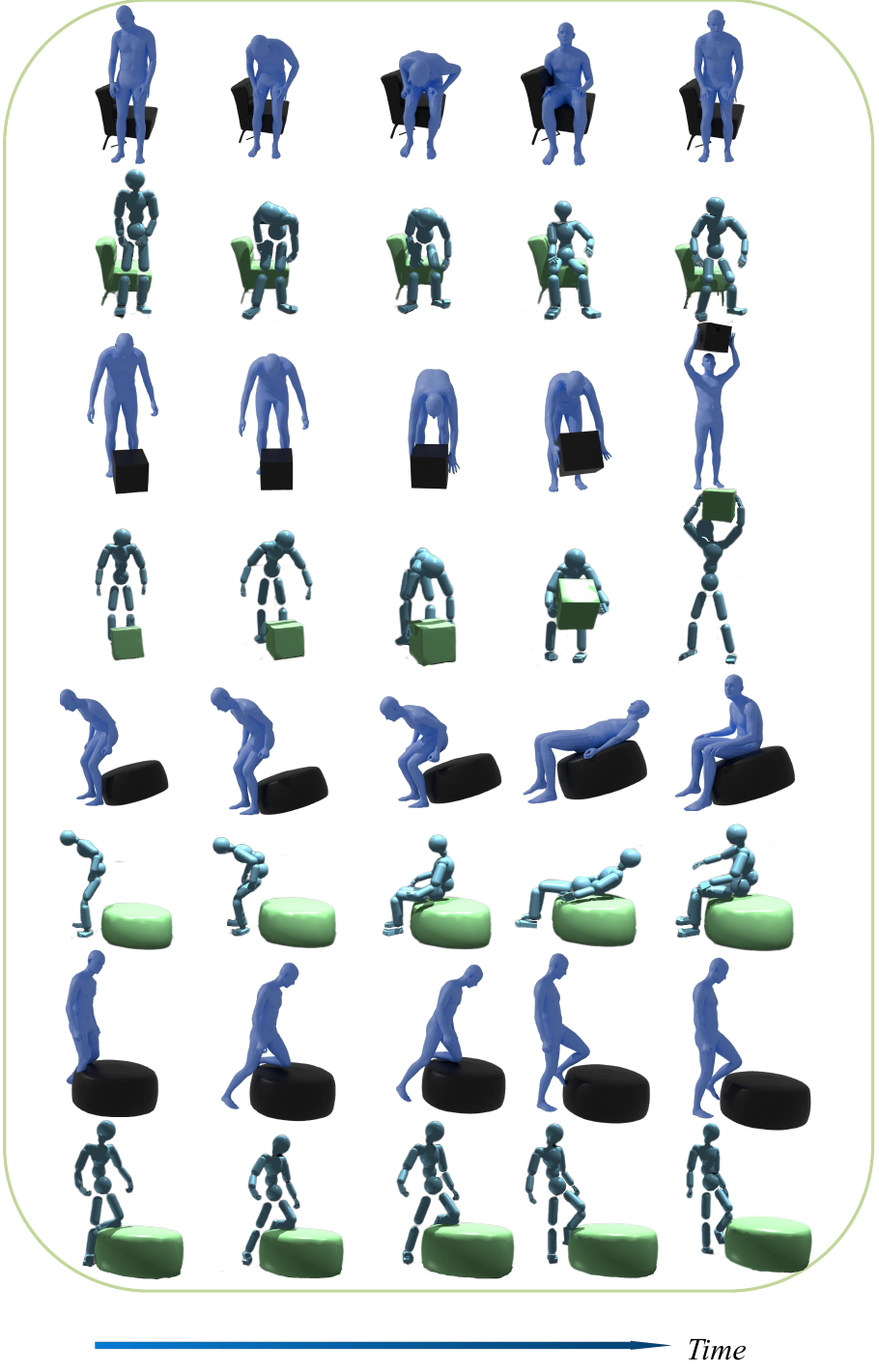}
    \caption{Qualitative results of the HOI simulation.}
    \label{fig:physic_sup}
\end{figure}

In this section, we provide additional visualizations of the HOI simulation results in Fig.~\ref{fig:physic_sup}. These examples further illustrate the overall motion quality, the temporal coherence of the simulated interactions, and the physical plausibility of the reconstructed human--object dynamics. They show that the simulated motions remain stable over time while preserving reasonable human--object coordination and interaction structure. The results also demonstrate that our simulation pipeline can maintain consistent motion patterns and realistic contact behaviors across a variety of interaction sequences.

%
%
\bibliographystyle{splncs04}
\bibliography{main}
\end{document}